\pdfoutput=1
\PassOptionsToPackage{table}{xcolor}
\documentclass[10pt]{article}

\usepackage[letterpaper,textwidth=5.5in,textheight=9in,top=1in,headheight=12pt,headsep=25pt,footskip=30pt]{geometry}
\usepackage{times}
\usepackage[T1]{fontenc}
\usepackage[utf8]{inputenc}
\usepackage{microtype}

\usepackage{amsmath}
\usepackage{amssymb}
\usepackage{amsfonts}
\usepackage{bm}

\usepackage{natbib}

\usepackage{titlesec}
\titleformat*{\section}{\large\bfseries}
\titleformat*{\subsection}{\normalsize\bfseries}
\titleformat*{\subsubsection}{\normalsize\bfseries}
\titlespacing*{\section}{0pt}{12pt}{6pt}
\titlespacing*{\subsection}{0pt}{9pt}{4pt}
\titlespacing*{\subsubsection}{0pt}{8pt}{3pt}

\usepackage{url}
\usepackage{enumitem}
\usepackage{booktabs}
\usepackage{tabularx}
\usepackage{xcolor}
\usepackage{multirow}
\usepackage{graphicx}
\usepackage{subcaption}
\usepackage{float}
\usepackage{wrapfig}
\usepackage{placeins}

\usepackage{tcolorbox}
\tcbuselibrary{listings,breakable}
\usepackage[colorlinks=true,linkcolor=blue!60!black,citecolor=blue!60!black,urlcolor=blue!60!black]{hyperref}

\renewenvironment{abstract}{%
  \vspace{0.3ex}%
  \centerline{\large\bfseries Abstract}%
  \vspace{0.5ex}%
  \begin{list}{}{\leftmargin=0.5in\rightmargin=0.5in\listparindent=0pt\itemindent=0pt\parsep=5.5pt}\item\relax
}{%
  \end{list}\vspace{1ex}%
}

\newcommand{\email}[1]{{\hypersetup{urlcolor=black}\href{mailto:#1}{\texttt{#1}}}}

\begin{document}

\begin{center}
  \hrule height 4pt
  \vskip 0.25in
  {\LARGE\bfseries TokenCast: Forecasting Token Consumption\\During LLM Agent Execution\par}
  \vskip 0.29in
  \hrule height 1pt
  \vskip 0.3in
  {\bfseries
  Chaoqian Ouyang\textsuperscript{1*}\quad
  Ling Yue\textsuperscript{2*}\quad
  Libin Zheng\textsuperscript{1\textdagger}\quad
  Hanghui Guo\textsuperscript{3}\quad
  Shengxiang Xu\textsuperscript{3}\\
  YiShu Wang\textsuperscript{3}\quad
  Ran Li\textsuperscript{4}\quad
  Jian Yin\textsuperscript{1}\quad
  Shaowu Pan\textsuperscript{2}\quad
  Shimin Di\textsuperscript{3\textdagger}\par}
  \vspace{2pt}
  \textsuperscript{1}Sun Yat-Sen University\quad
  \textsuperscript{2}Rensselaer Polytechnic Institute\quad
  \textsuperscript{3}Southeast University\\\quad
  \textsuperscript{4}Hong Kong University of Science and Technology\\
  \email{zhenglb6@mail.sysu.edu.cn}\quad
  \email{shimin.di@seu.edu.cn}
  \vspace{0.25in}
\end{center}

\begingroup
\renewcommand{\thefootnote}{\fnsymbol{footnote}}
\footnotetext[1]{Equal contribution.}
\footnotetext[2]{Corresponding authors.}
\renewcommand{\thefootnote}{}
\footnotetext{This work is conducted during the internship in Prof.\ Di's group.}
\endgroup
\setcounter{footnote}{0}

\begin{abstract}
When a large language model (LLM) agent executes the same task, token consumption can vary by over an order of magnitude across runs. The agent chooses its next steps based on tool feedback and intermediate results, while the growing context steadily inflates the input size of every subsequent call. The total consumption of a task is therefore hard to predict before execution and the prediction must be revised as the run unfolds. In this paper, we propose TokenCast, which learns a composable cost representation for each execution segment, recording its own consumption and the context growth it introduces. Composing adjacent segments yields a cumulative estimate that captures the extra input cost incurred when context from earlier segments is re-read by every later call. As execution unfolds, newly observed evidence refreshes the forecast, requiring no additional LLM calls and incurring a mean cumulative prediction time of 32.8\,ms per run on SWE-bench Verified. Across 4 task suites and 6 agent models, TokenCast's mean absolute error reduction against the strongest comparator averages 14.5\% over 96 evaluated combinations. In offline budget-control replay, TokenCast uses 21.3\% fewer tokens on average than a fixed-budget policy at matched trace completion. The code is available at \url{https://github.com/DEFENSE-SEU/TokenCast}.
\end{abstract}

\section{Introduction}

The applications of large language models (LLMs) are expanding from simple question answering to complex tasks such as software engineering and deep research. Completing these tasks typically relies on LLM-based agents that repeatedly plan, modify code, invoke tools, and verify results~\citep{DBLP:conf/iclr/YaoZYDSN023,DBLP:conf/nips/YangJWLYNP24,DBLP:conf/iclr/0001LSXTZPSLSTL25}. A given task may be resolved in a single edit or may require multiple rounds of retries before converging, and the dialogue and tool outputs generated in each round accumulate into the context of subsequent requests~\citep{DBLP:conf/nips/YangJWLYNP24,DBLP:journals/pacmse/XiaoGPX26,zhu2026tracelab}. The final token consumption of a task is therefore difficult to predict before execution completes, making it hard for users to forecast costs and plan usage~\citep{DBLP:journals/corr/abs-2604-22750}.

The challenge of predicting tokens lies in the fact that every action an agent takes can trigger new model calls, and consumption across steps is interdependent: the longer the context left by earlier steps, the larger the input to every subsequent call, causing consumption to accumulate and amplify~\citep{salim2026tokenomics,zhu2026tracelab,DBLP:journals/pacmse/XiaoGPX26}. The model's generative behavior further compounds the difficulty. Given the same input, the model may choose different courses of action, generate outputs of different lengths, and consequently undergo different numbers of verification or retry cycles. In practice, token consumption across different executions of the same task can differ by up to 30$\times$~\citep{DBLP:journals/corr/abs-2604-22750}, making accurate prediction challenging.
\begin{wrapfigure}{r}{0.58\textwidth}
  \centering
  \includegraphics[width=\linewidth]{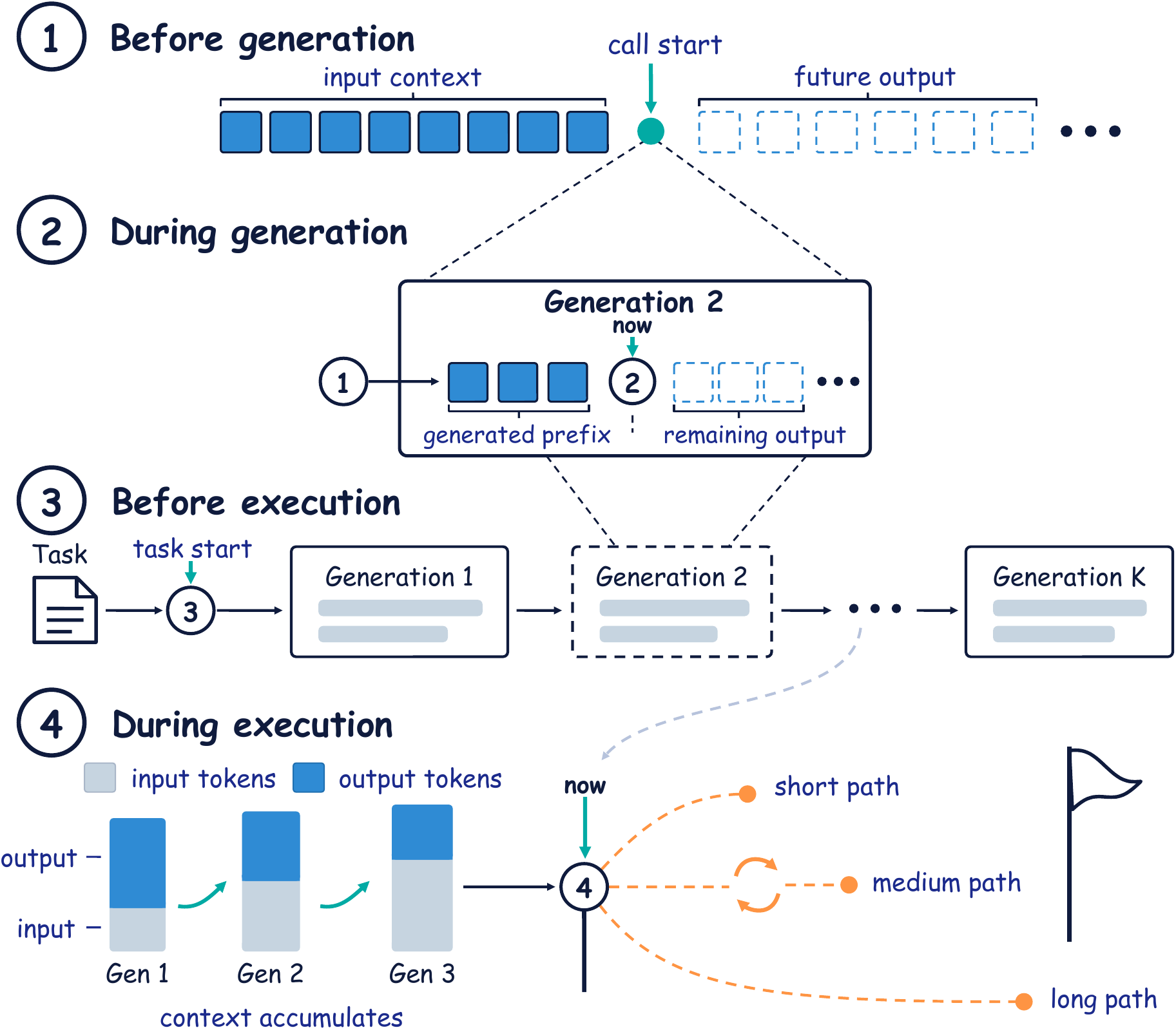}
\caption{Four token-consumption prediction settings. At the call level, the input context supports a forecast before generation (1), and the output prefix updates it during generation (2). At the task level, total consumption is forecast before execution (3), and remaining consumption is updated as calls complete and context accumulates (4).}
\label{fig:settings}
\end{wrapfigure}

Existing work can be organized by prediction scope and observation time. The scope ranges from a single response to an entire agent task; the forecast can be made before execution or updated during it. These dimensions give four settings (Figure~\ref{fig:settings}): response length before generation, remaining response length during generation, total task consumption before execution, and remaining task consumption as an agent runs. Prior methods address these settings with different targets and access assumptions~\citep{DBLP:conf/iclr/ShahoutMLJYM25,DBLP:journals/corr/abs-2602-11812,DBLP:journals/corr/abs-2604-22750}.

At the request level, prior work studies how to predict the response length of a single model call. Because the input is known before generation begins, these methods typically estimate length from input features for resource allocation~\citep{DBLP:conf/nips/JinW0W23,DBLP:journals/corr/abs-2404-08509,fu2024efficient,zheng2026scheduling}, while some also refine the estimate during generation using intermediate states or entropy statistics~\citep{DBLP:conf/iclr/ShahoutMLJYM25,DBLP:journals/corr/abs-2602-11812,DBLP:journals/corr/abs-2607-05316}.

At the agent-task level, prior work estimates total token consumption before execution, either from the task description or through the model's own cost assessment~\citep{DBLP:journals/corr/abs-2604-22750}. Multi-step LLM execution frameworks organize or schedule requests according to program structures, semantic dependencies, or workflow paths available before the corresponding requests are executed~\citep{DBLP:journals/corr/abs-2310-03714,DBLP:conf/osdi/LinHZ00CQ24,DBLP:conf/nips/ZhengYXS0YCKSGB24,DBLP:journals/corr/abs-2603-22206,yu2026pythia}. Open-ended agents present a different situation: their behavior depends on tool feedback, environment state, and intermediate results, and no complete dependency graph exists before execution begins~\citep{DBLP:conf/iclr/YaoZYDSN023,DBLP:conf/nips/YangJWLYNP24,DBLP:conf/iclr/0001LSXTZPSLSTL25}.

Multi-step forecasting has also been studied in structured workflows~\citep{DBLP:journals/corr/abs-2603-22206}. We focus on remaining provider-accounted input and output consumption in sequential agent execution. Each future request can bill retained context again, coupling the remaining cost to both future call count and input lengths. TokenCast updates this forecast from observed execution information without querying an LLM for the cost estimate.

\begin{figure}[!t]
\centering
\includegraphics[width=\textwidth]{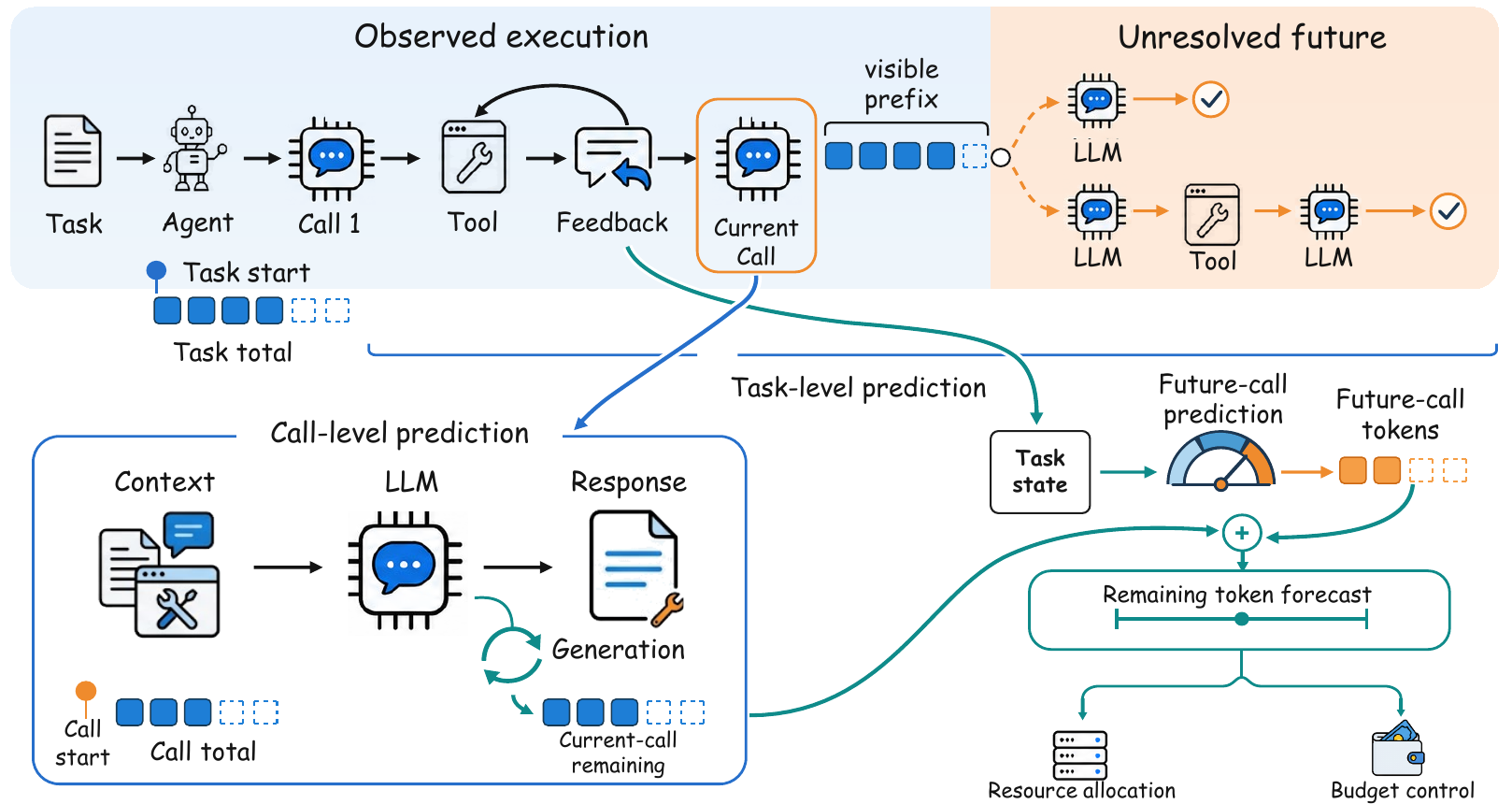}
\caption{Overview of TokenCast. Observed execution evidence supports call- and task-level forecasts, while the compositional path propagates predicted context growth into later-call costs.}
\label{fig:overview}
\end{figure}

Figure~\ref{fig:overview} summarizes TokenCast. It represents each execution segment by call count, net input-length change, and a cost residual. An exact composition identity exposes how context growth in one segment affects the cost baseline of later calls. The task predictor combines a direct forecast with a prefix--suffix forecast conditioned on a predicted boundary state, and refreshes its features as execution proceeds. Calibrated quantile models provide prediction intervals. The predictor makes no additional LLM calls.

Our contributions are as follows:
\begin{itemize}[leftmargin=*]
\item We introduce a segment-cost factorization and an exact composition identity that accounts for repeated input consumption across calls.

\item We use this identity in a staged prefix--suffix predictor that combines direct and compositional forecasts and updates from execution evidence without additional LLM calls.

\item Across four benchmarks and six agent models, TokenCast's MAE reduction averages 14.5\% across 96 comparisons with the strongest comparator in each. Budget-control replay saves 21.3\% of tokens at matched trace completion.
\end{itemize}

\section{Related Work}

\noindent\textbf{Request-level output length prediction.}
Early methods extract features from the prompt to estimate response length for memory reservation, batching, or shortest-job-first scheduling~\citep{DBLP:conf/nips/JinW0W23,DBLP:journals/corr/abs-2404-08509}. Prompt-only estimates remain highly uncertain, and prompt-conditioned response lengths can exhibit broad or heavy-tailed distributions~\citep{DBLP:journals/corr/abs-2505-16881,DBLP:journals/corr/abs-2604-07931}. Subsequent work therefore models more than a single point estimate: one line of work learns pairwise length orderings among requests to set scheduling priorities~\citep{fu2024efficient}, and another fits a heavy-tailed log-$t$ distribution that lets the scheduler trade off average and tail latency~\citep{zheng2026scheduling}. Once generation begins, the decoder's own intermediate states become available. Several recent studies show that mid-layer representations or entropy statistics can continuously refine the remaining-length estimate during decoding, narrowing the gap between the initial guess and the actual length~\citep{DBLP:conf/iclr/ShahoutMLJYM25,DBLP:journals/corr/abs-2607-05316,DBLP:journals/corr/abs-2602-11812}. These methods share a common set of assumptions: the prediction target is a single response, the prompt is known, and the system has access to the prompt or to model internals. In agent tasks, later requests do not exist until earlier actions and tool calls complete, so none of these assumptions holds.

\noindent\textbf{Task-level and multi-step prediction.}
Prior work has extended prediction to entire tasks. Self-Prediction has a coding agent inspect its environment before execution and estimate input, output, and total token consumption by stage~\citep{DBLP:journals/corr/abs-2604-22750}. DSPy~\citep{DBLP:journals/corr/abs-2310-03714}, Parrot~\citep{DBLP:conf/osdi/LinHZ00CQ24}, and SGLang~\citep{DBLP:conf/nips/ZhengYXS0YCKSGB24} represent multi-step LLM applications through program structures or semantic dependencies available to serving runtimes. Open-ended agents expose no such structure before execution.

Chimera predicts remaining workflow output with a CPU-based quantile random forest~\citep{DBLP:journals/corr/abs-2603-22206}. Pythia profiles historical traces to infer likely workflow paths and role-level output lengths~\citep{yu2026pythia}. TokenCast's segment identity addresses the repeated input cost induced by carried context, and its forecasts use no additional LLM calls.

Trace studies of deployed agents report extensive iterative review loops in multi-agent software pipelines, and long contexts with short outputs and heavy prefix reuse in real coding-agent sessions~\citep{salim2026tokenomics,zhu2026tracelab}. Broader analyses organize token use and efficiency across single-agent, multi-agent, and agent-ecosystem settings~\citep{chen2026token}. These studies characterize consumption after the fact and do not forecast it for a running task.

\section{Method}
\label{sec:method}

\subsection{Problem Formulation}
\label{sec:problem_formulation}

An agent executes a task through a sequence of LLM calls interleaved with tool interactions. Let $C_k$ denote the provider-accounted input and output token consumption of call $k$. A task terminates upon completion, failure, or when an execution limit is reached. For a task with $K$ calls, its total consumption is $T=\sum_{j=1}^{K}C_j$. After call $k$ completes, the confirmed consumption is $S_k=\sum_{j=1}^{k}C_j$, and the remaining consumption is $R_k=T-S_k$. Each forecast uses only the task and execution information available at its prediction point.

We consider four prediction settings along the execution trajectory.
\begin{itemize}[leftmargin=*]
\item \textbf{Task Start} predicts the total consumption $T$ before execution begins.
\item \textbf{Call Start} predicts the consumption $C_k$ of the current call after its request has been assembled and before any output token is generated.
\item \textbf{In-call Update} updates the prediction of $C_k$ as the generated prefix becomes available.
\item \textbf{Task Update} predicts the remaining consumption $R_k$ after call $k$ completes, yielding an updated forecast of the task total, $\widehat{T}_k=S_k+\widehat{R}_k$.
\end{itemize}

\subsection{Call--Task Forecasting}
\label{sec:call_task_forecasting}

TokenCast represents segment cost relative to its starting input length. The next segment inherits the context produced by the preceding one, yielding an exact composition rule for adjacent segments.

\noindent\textbf{Segment representation.}
A contiguous block of one or more calls forms a segment. Let segment $A$ start with input length $L_A$, span $n_A$ calls, and consume $C_A$ tokens in total. Its representation is $\phi_A=(n_A,\,g_A,\,b_A)$, where $g_A$ is the net change in input length across the segment and $b_A=C_A - n_A L_A$ is the residual after subtracting the starting-input baseline $n_A L_A$, covering generation and within-segment context growth. The next segment starts with input length $L_A+g_A$. Reasoning tokens billed by the provider contribute to $b_A$; when they do not persist in the conversation context, they do not contribute to the context change $g_A$.

When segment $B$ immediately follows $A$, the combined representation is
\begin{equation}
    \phi_{A\circ B}
    =
    \bigl(
    n_A+n_B,\;
    g_A+g_B,\;
    b_A+b_B+n_B\,g_A
    \bigr).
    \label{eq:segment_composition}
\end{equation}
This identity follows from the definitions and the boundary condition $L_B = L_A + g_A$.
The third term $n_B\,g_A$ arises from realigning the baseline: $b_B$ is defined relative to $B$'s actual starting point $L_A+g_A$, whereas the combined residual is relative to $L_A$. The difference on each call is exactly $g_A$, and $n_B$ calls accumulate to $n_B\,g_A$.

\noindent\textbf{Forecasting with composition.}
The decomposition converts aggregate remaining-cost prediction into three sub-problems: the prefix segment's representation, the boundary state, and the suffix segment's representation. TokenCast fits LightGBM models for these predictions and updates after completed calls. Appendix~\ref{app:base_predictor} compares alternative base predictors. Input features are drawn from the task and execution information available at the prediction point. Appendix~\ref{app:execution_evidence} details these features and when each becomes available.

Call-level forecasts predict the current call's cost from the features visible at Call Start or In-call Update. Task-level forecasts maintain two paths. The direct path predicts remaining total cost as a single target. The compositional path separates the current or next segment from the subsequent suffix. Their cost representations are predicted separately and combined using Eq.~(\ref{eq:segment_composition}).

The compositional path predicts the prefix representation $(\widehat{n}_A,\widehat{g}_A,\widehat{b}_A)$ and its ending state. The predicted context change sets the suffix input baseline, while the predicted ending state and features visible at the original prediction point condition the suffix model. The suffix representation is converted to a remaining-cost forecast through Eq.~(\ref{eq:segment_composition}).

\noindent\textbf{Updating forecasts.}
Each completed step during execution produces new observations, and TokenCast refreshes its forecasts accordingly. At Call Start, the current request has been assembled and the actual input length is known, so the call-level forecast can be based on it directly. At In-call Update, the committed generated prefix and streaming timing supply additional features for the current-call forecast. At Task Update, the completed call supplies confirmed cumulative cost $S_k$ and any completed tool outcomes. The next request's input length remains predicted until that request is assembled. TokenCast uses the available evidence to re-predict $\widehat{R}_k$ and reports $\widehat{T}_k=S_k+\widehat{R}_k$.

\subsection{Compositional Learning}
\label{sec:compositional_learning}

\noindent\textbf{Staged fitting.}
TokenCast fits the direct, prefix, and suffix predictors as separate LightGBM models using labels extracted from completed traces. The direct path predicts remaining total cost. The compositional path predicts the prefix and suffix representations together with the prefix-ending boundary variables. Prefix context-change and suffix call-count errors can affect downstream cost through the composition identity, motivating the following weights on their local absolute losses:
\begin{equation}
    \ell_g
    =
    n_B\left|\widehat{g}_A-g_A\right|,
    \qquad
    \ell_n
    =
    L_B\left|\widehat{n}_B-n_B\right|.
    \label{eq:local_cost_weighting}
\end{equation}

\noindent\textbf{Cross-fitting.}
The suffix model is trained on predicted prefix boundaries. Tasks are partitioned into $F$ folds, and each fold's prefix predictions are generated by models trained on the remaining folds. The predicted context change sets the suffix input baseline, so its residual label is recomputed as $b_B^{\mathrm{train}}=C_B-n_B\widehat{L}_B$, where $\widehat{L}_B=L_A+\widehat{g}_A^{\mathrm{oof}}$. The loss weights in Eq.~(\ref{eq:local_cost_weighting}) are motivated by the composition identity; Appendix~\ref{app:theoretical_analysis} distinguishes this motivation from the error decomposition under the rebased suffix training label.

\noindent\textbf{Correction model.}
After the direct, prefix, and suffix models are fixed, a correction model is trained on out-of-fold outputs of the complete forecasting pipeline:
\begin{equation}
    \psi^{*}
    =
    \operatorname*{arg\,min}_{\psi}
    \sum_i
    \left|
    y_i
    -
    \widehat{C}_{\mathrm{comp},i}
    -
    h_{\psi}(\mathbf{q}_i)
    \right|.
    \label{eq:task_correction}
\end{equation}
The input $\mathbf{q}_i$ contains the visible features, the direct and compositional forecasts, their difference, and the predicted boundary variables. The final task-level forecast is $\widehat{C}_{\mathrm{comp}}+h_{\psi}(\mathbf{q})$.

\noindent\textbf{Prediction intervals.}
Separate LightGBM quantile models produce the 0.05 and 0.95 endpoints at each prediction point. The endpoints are widened symmetrically by a quantile of interval residuals on held-out calibration tasks and are bounded below by consumption already confirmed within the prediction scope. Appendix~\ref{app:interval_calibration} gives the procedure.

\section{Experiments}
\label{sec:experiments}

\begin{table}[t]
\centering
\small
\caption{MAE $\downarrow$ in tokens at four prediction points. Norm.\ Avg.\ is the macro-average of MAE normalized at each prediction point by the MAE of the corresponding history-median predictor. $k$ denotes thousands of tokens. Best and second-best results are in bold and underlined, respectively. Marks are assigned using unrounded values.}
\label{tab:main}
\setlength{\tabcolsep}{4.5pt}
\begin{tabular}{llrrrrr}
\toprule
Agent LLM & Prediction point & TokenCast (Ours) & TRAIL & EGTP & TIE & Self-Pred. \\
\midrule
\multicolumn{7}{c}{\textbf{\textit{SWE-bench Verified}}} \\
\midrule
\multirow{4}{*}{\textbf{GPT-5.4}}
& Task Start     & \textbf{144.0k} & 165.3k & 160.9k & 157.0k & \underline{152.0k} \\
& Call Start     & \textbf{64.5} & 71.1   & 70.3   & \underline{69.5} & 70.9 \\
& In-call Update & \textbf{38.9} & 78.9   & \underline{74.6} & 77.4 & 80.2 \\
& Task Update    & \textbf{80.0k}  & \underline{115.0k} & 117.0k & 119.0k & 126.0k \\
\rowcolor{gray!12}
& \textbf{Norm. Avg.} & \textbf{0.69} & 0.94 & \underline{0.92} & 0.92 & 0.94 \\
\cmidrule(lr){1-7}

\multirow{4}{*}{\textbf{Qwen3.8-27B}}
& Task Start     & \textbf{192.0k} & 232.0k & \underline{209.0k} & 215.0k & 221.0k \\
& Call Start     & \textbf{87.0}   & 109.0  & 102.9 & \underline{99.0} & 105.5 \\
& In-call Update & \textbf{81.9} & 120.0  & 116.0   & 111.4 & \underline{108.9} \\
& Task Update    & \textbf{124.0k} & 176.0k & 171.0k  & \underline{164.0k} & 173.0k \\
\rowcolor{gray!12}
& \textbf{Norm. Avg.} & \textbf{0.65} & 0.86 & 0.81 & \underline{0.79} & 0.82 \\

\midrule
\multicolumn{7}{c}{\textbf{\textit{Search-R1}}} \\
\midrule
\multirow{4}{*}{\textbf{GPT-5.4}}
& Task Start     & \textbf{34.2k} & 37.8k & 37.6k & \underline{36.4k} & 36.9k \\
& Call Start     & \underline{34.5} & 37.3 & 36.5 & \textbf{34.0} & 39.8 \\
& In-call Update & \textbf{31.3} & 39.4 & 38.5 & \underline{36.8} & 39.9 \\
& Task Update    & \textbf{17.6k} & \underline{23.1k} & 23.7k & 23.3k & 24.2k \\
\rowcolor{gray!12}
& \textbf{Norm. Avg.} & \textbf{0.75} & 0.89 & 0.88 & \underline{0.85} & 0.91 \\
\cmidrule(lr){1-7}

\multirow{4}{*}{\textbf{Qwen3.8-27B}}
& Task Start     & \textbf{39.0k} & 53.5k & 49.8k & \underline{47.6k} & 51.0k \\
& Call Start     & \textbf{56.7} & \underline{65.9} & 71.4 & 67.7 & 76.1 \\
& In-call Update & \textbf{57.0} & 84.6 & 80.7 & \underline{75.5} & 83.2 \\
& Task Update    & \textbf{24.0k} & 39.5k & \underline{34.0k} & 35.4k & 38.4k \\
\rowcolor{gray!12}
& \textbf{Norm. Avg.} & \textbf{0.59} & 0.83 & 0.79 & \underline{0.77} & 0.84 \\
\bottomrule
\end{tabular}
\end{table}

\subsection{Experimental Setup}
\label{sec:experimental_setup}

\noindent\textbf{Tasks and execution traces.}
We evaluate TokenCast on SWE-bench Verified~\citep{DBLP:conf/iclr/JimenezYWYPPN24, chowdhury2024swebenchverified}, Search-R1~\citep{DBLP:journals/corr/abs-2503-09516}, MMLU-Pro~\citep{DBLP:conf/nips/WangMZNCGRAHJLK24}, and LongBench-v2~\citep{DBLP:conf/acl/BaiTZ0WLCX0D0L25}, covering software engineering, retrieval-based question answering, knowledge-based reasoning, and long-context understanding. We collect 11,712 execution traces from 240 benchmark tasks with six agent LLMs: GPT-5.4~\citep{openai2026gpt54}, Claude Opus 4.6~\citep{anthropic2026claudeopus46}, Gemini 3.1 Pro~\citep{googledeepmind2026gemini31pro}, DeepSeek-V4-Pro~\citep{deepseekai2026deepseekv4}, Qwen3.8-27B~\citep{qwen38}, and Llama-3.2-3B-Instruct~\citep{meta2024llama32}. For reproducibility, we specify \textsl{gpt-5.4-2026-03-05} for the GPT-5.4 API and \textsl{qwen3.8-27b-20260815} for the self-hosted Qwen checkpoint; Table~\ref{tab:agent_llms} lists model access and reasoning configurations. Traces are collected using DeepSeek Harness~\citep{deepseekharness} and OpenHands~\citep{DBLP:conf/iclr/0001LSXTZPSLSTL25}. Repeated executions support the analysis of run-to-run variation, with additional repeats for anchor tasks. Table~\ref{tab:trace_design} specifies the collection design for each benchmark and harness. All runs of the same task remain in one partition. Training, validation, calibration, and test tasks are separated as described in Appendix~\ref{app:experimental_details}. Generalization experiments also use independently released trajectories (Appendix~\ref{app:collection}).

\noindent\textbf{Baselines.}
We compare TokenCast with three output-length predictors, TRAIL~\citep{DBLP:conf/iclr/ShahoutMLJYM25}, EGTP~\citep{DBLP:journals/corr/abs-2602-11812}, and TIE~\citep{zheng2026scheduling}, and the agent-level consumption estimator Self-Prediction~\citep{DBLP:journals/corr/abs-2604-22750}. We adapt these methods to the prediction targets and observations available at the four prediction points. At Task Start, Self-Prediction inspects the task environment before estimating total consumption. At the other prediction points, it uses the observed execution prefix. Appendix~\ref{app:baselines} details each adaptation. Appendix~\ref{app:comp_vs_direct} compares direct regression, compositional forecasting, their average, and the full correction pipeline; Appendix~\ref{app:segment_ablation} evaluates the segment representation and composition procedure.

\noindent\textbf{Metrics and implementation.}
We report mean absolute error (MAE) and weighted absolute percentage
error (WAPE) at the four prediction points. At Call Start, the target includes the assembled request's known input tokens. For 90\% prediction intervals, we report empirical coverage, mean width, and mean interval score (MIS)~\citep{gneiting2007strictly}. Tasks receive equal weight, with that weight distributed across their runs and evaluated checkpoints. Cross-configuration summaries divide MAE by that of a history-median predictor, which outputs the median training target for the same benchmark, agent LLM, and prediction point. The local platform provides eight NVIDIA A100 GPUs. Internal-state baselines use the agent model when accessible and a proxy for API-based models. Appendix~\ref{app:experimental_details} gives metrics, data partitions, training settings, and timing procedures.

\subsection{Experimental Results and Analysis}
\label{sec:results}
\subsubsection{Main Results}
\label{sec:main_results}

\begin{figure}[t]
\centering
\includegraphics[width=1.0\textwidth]{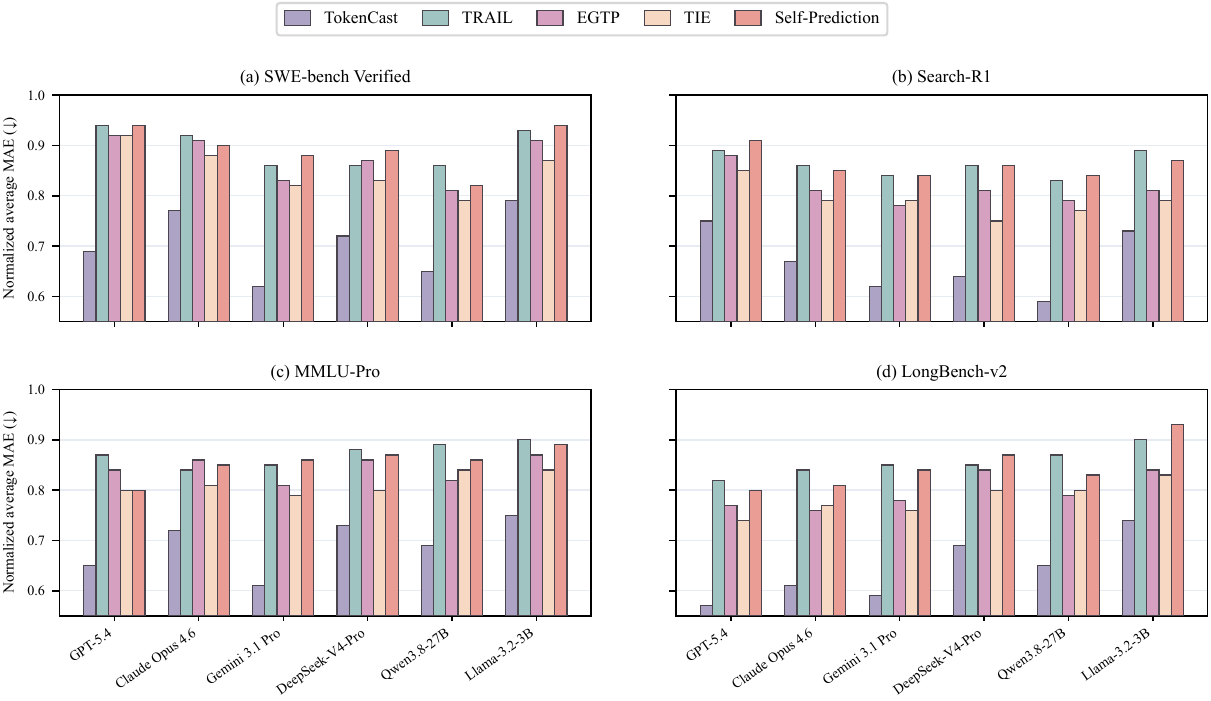}
\caption{Normalized average MAE across four benchmarks and six agent LLMs. Each panel corresponds to one benchmark, and each bar group corresponds to one agent LLM. For each method, MAE is normalized by the corresponding history-median MAE at each prediction point and macro-averaged over the four prediction points. Lower is better.}
\label{fig:benchmark_accuracy}

\vspace{0.8em}
\captionof{table}{Execution traces per benchmark and harness, over six agent LLMs.}
\label{tab:trace_design}
\small
\setlength{\tabcolsep}{4pt}
\begin{tabular}{llrrrrr}
\toprule
Benchmark & Harness & Tasks & Runs & Anchors & Anchor runs & Traces \\
\midrule
SWE-bench Verified & DeepSeek Harness & 144 & 2 & 24 & 8 & 2,592 \\
SWE-bench Verified & OpenHands & 144 & 2 & 24 & 8 & 2,592 \\
Search-R1 & DeepSeek Harness & 32 & 4 & 8 & 16 & 1,344 \\
Search-R1 & OpenHands & 32 & 4 & 8 & 16 & 1,344 \\
MMLU-Pro & DeepSeek Harness & 32 & 4 & 8 & 8 & 960 \\
MMLU-Pro & OpenHands & 32 & 4 & 8 & 8 & 960 \\
LongBench-v2 & DeepSeek Harness & 32 & 4 & 8 & 8 & 960 \\
LongBench-v2 & OpenHands & 32 & 4 & 8 & 8 & 960 \\
\midrule
Total & & & & & & 11,712 \\
\bottomrule
\end{tabular}
\end{figure}

Table~\ref{tab:main} shows SWE-bench Verified and Search-R1 results for GPT-5.4 and Qwen3.8-27B; Figure~\ref{fig:benchmark_accuracy} covers all four benchmarks and six agent LLMs. On SWE-bench Verified with GPT-5.4, TokenCast reduces MAE relative to the strongest comparator by 47.9\% at In-call Update, from EGTP's 74.6 to 38.9 tokens, and by 30.4\% at Task Update, from TRAIL's 115k to 80k tokens. At Task Start, the strongest comparator is Self-Prediction, and the reduction is 5.3\%, from 152k to 144k tokens. Across the 96 benchmark--model--prediction-point combinations in Appendix~\ref{app:benchmark_results}, its MAE reduction against the lowest comparator MAE averages 14.5\%. The averages are $-2.2\%$ at Task Start, $1.9\%$ at Call Start, $30.4\%$ at In-call Update, and $27.8\%$ at Task Update. TokenCast trails the strongest comparator in 24 combinations: 15 at Task Start and nine at Call Start. None of these losses occurs at In-call Update or Task Update, where execution evidence accumulates.

\noindent\textbf{Error relative to consumption.}
WAPE complements MAE by expressing absolute error relative to mean target consumption. For GPT-5.4 at Task Start, TokenCast's MAE of 32.2k tokens on LongBench-v2 corresponds to a WAPE of 4.2\%, whereas its MAE of 8.0k on MMLU-Pro corresponds to 15.8\%. Their mean target consumptions are 771.2k and 50.6k tokens, respectively. Table~\ref{tab:wape} in Appendix~\ref{app:wape} reports the full WAPE results.

\subsubsection{Prediction Reliability}
\label{sec:prediction_reliability}

\noindent\textbf{Run-to-run variation.} Repeated GPT-5.4 executions of the same task show substantial consumption spread across all four benchmarks. Figure~\ref{fig:repeated_runs} in Appendix~\ref{app:prediction_reliability} shows the distributions for 48 anchor tasks and gives the repetition counts per benchmark.

\noindent\textbf{Interval reliability.} On the anchor tasks, TokenCast's calibrated intervals reduce MIS relative to Self-Prediction's native intervals at all four prediction points, by 32.0\% on average. At Task Start, TokenCast covers 82.0\% of outcomes against a nominal 90\% level, while Self-Prediction covers 52.7\%. Table~\ref{tab:repeated_runs} reports interval width, coverage, and MIS at every prediction point.

\subsubsection{Generalization}
\label{sec:generalization}

\noindent\textbf{Unseen task types.} On independently released LiveClawBench trajectories~\citep{long2026liveclawbench}, zero-shot leave-one-domain-out transfer yields MAE ratios of 1.31 at Call Start and 1.47 at Task Update relative to Self-Prediction. With 20 target-domain tasks for adaptation, the ratios fall to 0.82 and 0.85. Appendix~\ref{app:unseen_task_types} gives the protocol and intermediate results.

\noindent\textbf{Unseen agent LLMs.} At Call Start, transfer to a held-out agent LLM yields 1.02 times the target history-median MAE without target-model tasks, improving to 0.95 with 20 tasks. With 3--10 target tasks, transfer outperforms target-only training. Appendix~\ref{app:unseen_llms} gives the protocol and a separate remaining-output-token Task Update analysis; other generalization results appear in Appendix~\ref{app:generalization}.

\subsection{Case Study}
\label{sec:budget_control}

TokenCast's forecasts are revised as execution unfolds. In a code-repair run, a verification failure raises the task-total forecast, which decreases after a compatibility workaround passes the reported assertions. In long-context QA, file operations after answer generation consume further tokens and raise the forecast. Both traces are in Appendix~\ref{app:case_studies}. We next examine token-budget control driven by these online forecasts on SWE-bench Verified.

\begin{figure}[t]
\centering
\includegraphics[width=\textwidth]{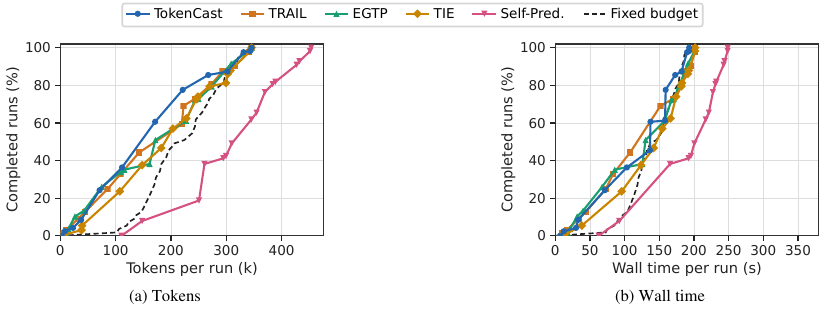}
\caption{Trace completion under stopping limits on SWE-bench Verified. Panel (a) plots trace completion against mean tokens per run, and panel (b) plots it against replay-accounted wall time per run. Prediction overhead is included. The dashed curve is the fixed-budget baseline.}
\label{fig:budget_control}

\vspace{0.8em}
\small
\captionof{table}{90\% prediction intervals over repeated executions of anchor tasks. TokenCast uses held-out calibration; Self-Prediction uses its reported 5th and 95th percentiles.}
\label{tab:repeated_runs}
\begin{tabular}{lrrrrrr}
\toprule
& \multicolumn{3}{c}{TokenCast}
& \multicolumn{3}{c}{Self-Prediction} \\
\cmidrule(lr){2-4}\cmidrule(lr){5-7}
Prediction point & Coverage (\%) & Width & MIS $\downarrow$ & Coverage (\%) & Width & MIS $\downarrow$ \\
\midrule
Task Start & 82.0 & 752.5k & \textbf{1547.9k} & 52.7 & 354.3k & 3334.6k \\
Call Start & 89.8 & 315 & \textbf{549} & 73.0 & 223 & 638 \\
In-call Update & 92.8 & 319 & \textbf{576} & 75.2 & 237 & 732 \\
Task Update & 91.2 & 379.1k & \textbf{574.3k} & 77.0 & 232.0k & 941.6k \\
\bottomrule
\end{tabular}
\end{figure}

\noindent\textbf{Budget control.} On 288 GPT-5.4 runs from 144 SWE-bench Verified tasks, Figure~\ref{fig:budget_control} shows that TokenCast uses 21.3\% fewer tokens on average across seven replay budgets while matching fixed-budget trace completion at every budget. A run is trace-complete when it reaches its recorded terminal state; the replay does not measure task resolution. Prediction costs are included. Appendix~\ref{app:budget_control} gives the stopping rules, per-budget results, and overhead (Table~\ref{tab:prediction_overhead}).

\begin{figure}[t]
\centering
\includegraphics[width=\linewidth]{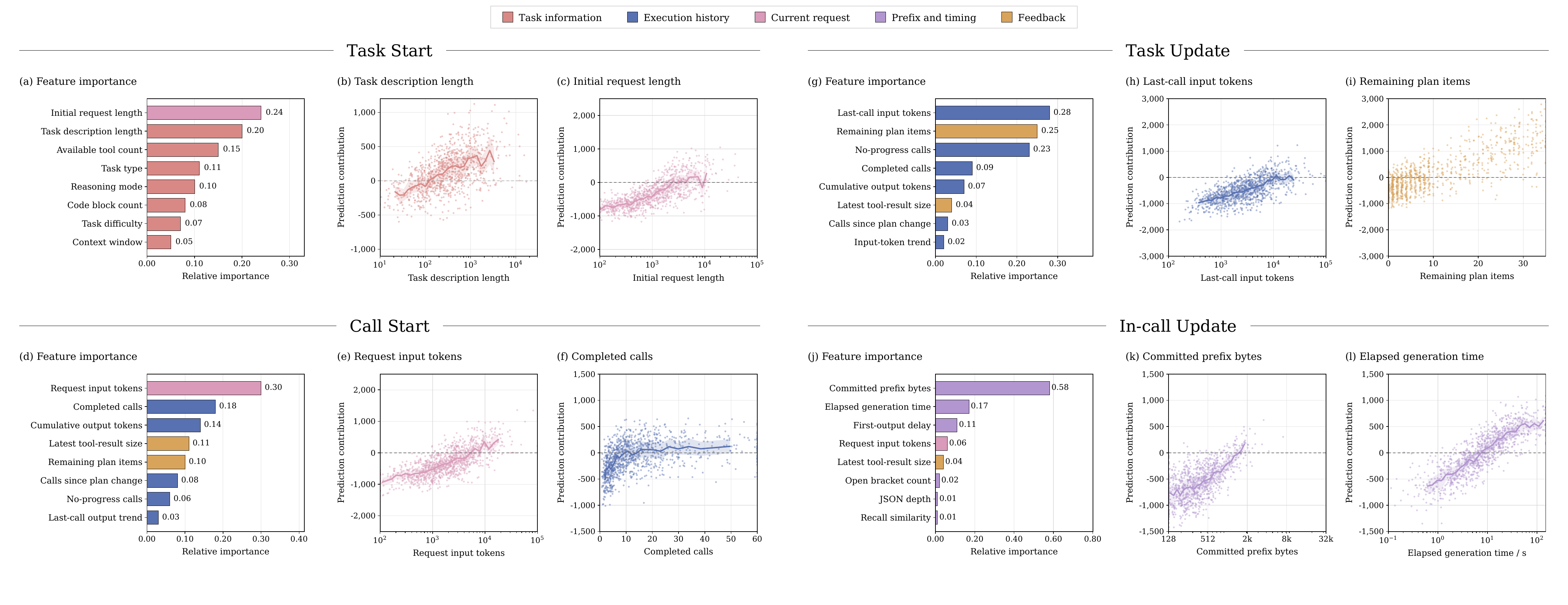}
\caption{Feature analysis at the four prediction points. For each point, the panels show the relative importance of selected features and contribution patterns for two representative features. Colors indicate the feature groups defined in Appendix~\ref{app:execution_evidence}.}
\label{fig:source_ablation}
\end{figure}

\subsection{Ablation Study}
\label{sec:ablation}

\noindent\textbf{Feature analysis.} Figure~\ref{fig:source_ablation} shows that the useful evidence changes over a run. Request length leads at Task Start and Call Start, with relative importance of 0.24 and 0.30. During generation, committed prefix length dominates at 0.58. After a call completes, last-call input tokens, remaining plan items, and calls without progress carry similar importance at 0.28, 0.25, and 0.23.

\noindent\textbf{Segment representation.} The segment triple improves both task-level and call-level forecasts. Replacing it with total token count, input length, and output length raises their normalized MAEs from 0.71 to 0.88 and from 0.68 to 0.82. The no-composition variant has a normalized average MAE of 0.78 and call-level MAE of 0.74, compared with 0.69 and 0.68 for the full method. Within the triple, omitting context growth yields 0.76, while omitting the residual yields 0.73.

\noindent\textbf{Forecasting strategy.} Before execution, direct regression has lower normalized MAE than composition, 0.82 versus 0.87. The ranking reverses at Task Update after calls have completed: composition reaches 0.66 and direct regression 0.72. Averaging the two forecasts reduces Task Update MAE to 0.64, and the full correction model reaches 0.62 while raising interval coverage from 88.1\% under direct regression to 90.6\%. Appendix~\ref{app:comp_vs_direct} gives the full comparison.

\noindent\textbf{Pipeline components.} The largest degradation comes from dropping both cross-fitting and cost weighting. Normalized average MAE rises from 0.69 to 0.78, while 90\% interval coverage falls from 90.6\% to 87.4\%. Dropping cross-fitting alone gives 0.73 MAE, and dropping cost weighting alone gives 0.72. The correction model and predicted boundary variables also contribute: their separate ablations give 0.74 and 0.72 MAE. Among six base predictors, LightGBM gives the lowest error and the highest coverage with 0.8\,ms model time. Appendix~\ref{app:component_ablation} reports the complete component and base-predictor results in Tables~\ref{tab:component_ablation} and~\ref{tab:base_predictor}.

\noindent\textbf{Update frequency.} Updating after every call takes 32.8\,ms of mean cumulative prediction time and makes 19.7 forecasts per run. Updating every three calls cuts the time to 11.9\,ms and the forecast count to 6.9, and a single forecast at Task Start takes 2.1\,ms. These full-pipeline times include feature extraction and model inference, so every-call updating adds under 0.03\% to the 129\,s median wall time of a GPT-5.4 run. Appendix~\ref{app:update_frequency} reports the full frequency sweep with per-run standard deviations for all four update intervals.

\section{Conclusion}
TokenCast addresses the problem of forecasting token consumption during LLM agent execution, where context accumulation causes each call's cost to depend on the entire preceding trajectory. The method represents each execution segment with a composable cost triple that separates the starting-input baseline from incremental consumption, and composes adjacent segments to propagate context growth into downstream cost estimates. Across four benchmarks and six agent LLMs, TokenCast's MAE reduction against the strongest comparator averages 14.5\% over 96 evaluated combinations and, in offline budget-control replay, it matches a fixed budget's trace-completion rate while consuming 21.3\% fewer tokens. Its mean cumulative task-level prediction time is 32.8\,ms per run on SWE-bench Verified. Because the predictors are learned from recorded traces, transfer to a new task domain or agent LLM improves with a small set of target tasks from the new setting. This work provides a lightweight prediction layer for agent token consumption, and we leave the integration of these forecasts into a runtime decision framework that actively manages execution under budget constraints as a direction for future work.

\clearpage

\section*{Ethics Statement}
Our experiments use public benchmarks and recorded agent trajectories to study token consumption in LLM agents. The study involves no human subjects or collection of private user data. Models and datasets are used under their respective licenses and terms of use.

\section*{Reproducibility Statement}
Code associated with this study is linked at \url{https://github.com/DEFENSE-SEU/TokenCast}. Section~\ref{sec:method} presents the forecasting formulation and training objective, with implementation details in Appendix~\ref{app:method_details}. Appendix~\ref{app:experimental_details} documents benchmark sampling, trajectory collection, agent and harness configurations, data splits, baseline implementations, and evaluation metrics. The budget-control replay protocol and complete Self-Prediction prompt are provided in Appendices~\ref{app:budget_control} and~\ref{app:self_prediction_prompt}, respectively.

\section*{AI Use Statement}
During manuscript preparation, large language models were used solely as general-purpose writing assistants for grammar checking, word refinement, and improving clarity. LLMs did not contribute to the research ideation, methodological design, or experimental execution. All suggestions produced by the LLMs were reviewed, edited, and vetted by the authors, who take full responsibility for the entire content of the paper.

\bibliographystyle{plainnat}
\bibliography{main}

\clearpage

\appendix
\setcounter{topnumber}{2}
\setcounter{bottomnumber}{2}
\setcounter{totalnumber}{3}
\renewcommand{\topfraction}{0.9}
\renewcommand{\bottomfraction}{0.6}
\renewcommand{\textfraction}{0.1}
\renewcommand{\floatpagefraction}{0.75}
\makeatletter
\setlength{\@fptop}{0pt}
\setlength{\@fpsep}{12pt plus 1fil}
\setlength{\@fpbot}{0pt plus 1fil}
\makeatother
\section*{Appendix}

\FloatBarrier
\section{Comparison with Related Methods}
\label{app:related_work}

Table~\ref{tab:related_work} compares prediction targets, evidence, and update points. TokenCast forecasts call and remaining-task consumption from execution features and a segment-cost representation.

\begin{table}[htbp]
\centering
\caption{Prediction targets and estimators in related work. Appendix~\ref{app:baselines} describes our adaptations.}
\label{tab:related_work}
\small
\begin{tabularx}{\linewidth}{@{}l p{0.27\linewidth} X p{0.18\linewidth}@{}}
\toprule
Work & Prediction target & Evidence and estimator & Update point \\
\midrule
TRAIL & Response length & Last-token hidden state and length bins & Before generation \\
EGTP & Response and remaining-response length & Hidden states and entropy pooling & Pre/mid generation \\
TIE & Response length distribution & Text embedding and log-$t$ model & Before generation \\
Chimera & Remaining workflow output & Prompt and workflow features; quantile forest & Workflow request \\
Pythia & Workflow path and role output length & Historical trace profiles & Incoming request \\
Self-Prediction & Total task tokens & Agent inspection of task environment & Before execution \\
\textbf{TokenCast} & Provider-accounted call and remaining-task tokens & Execution features; direct and compositional predictors & Four prediction points \\
\bottomrule
\end{tabularx}
\end{table}

\FloatBarrier
\section{Method Details}
\label{app:method_details}

\FloatBarrier
\subsection{Execution Evidence}
\label{app:execution_evidence}

The observation record $\mathbf{v}$ contains only information available at the prediction point. Token usage comes from provider records, and a logical call aggregates its initial request and any retries. The input length of an assembled request is measured directly; a future request's input length remains predicted until assembly. The visible prefix contains streamed text and tool-call arguments committed so far, with a checkpoint every 128 UTF-8 bytes. Unavailable fields are masked. Prediction-point observations remain separate from the prefix-ending predictions passed to the suffix model.

\noindent\textbf{Task representation.}
The task representation $\mathbf{d}$ uses TF-IDF features over word unigrams and bigrams and character 3- to 5-grams, with 12,000 terms for each representation. When a benchmark provides a task-difficulty label, the label is prepended to the task statement before encoding.

\noindent\textbf{Action types.}
Reads include file reads, glob matches, and searches. Edits write or replace file content. Tracking updates the to-do list. A shell command is labeled as a test when it invokes pytest, tox, nox, or unittest, or refers to a test directory, and other shell commands are labeled as runs. An action is marked as failed when a tool reports an error or a shell command exits with a nonzero status. The stage of a call is determined by its most recent action.

\noindent\textbf{Execution history.}
The progress record contains the numbers of planned, completed, and remaining to-do items, together with the number of calls since the list was last updated. The consumption history contains the input, output, and reasoning usage of the latest completed call, their changes from the preceding call, their mean, median, and least-squares trends over the last five and ten calls, and running totals of calls, model requests, tool actions, and confirmed tokens. Task Update additionally records the numbers of failed requests and tool errors in the latest call, together with the numbers of consecutive calls that failed, required a retry, or made no progress. Progress is defined as an edit, test, tracking update, or completed plan item. A window statistic is masked when the required observations are unavailable.

\noindent\textbf{Recent tool actions.}
The $w$ most recent tool actions fill the slots in recency order. Each slot records the action type, status, result size, calls since execution, and the reported result: an exit code, the number of lines returned by a read, the number of matches returned by a search, or the passed and failed counts of a test run. We select $w$ per fold from $\{3,5,8\}$ using the validation tasks.

\noindent\textbf{Generated prefix.}
The lowercased prefix is hashed by words and word pairs into 256 signed buckets with weight $1+\log n$ for $n$ occurrences, and its last line is hashed into 64 buckets. A single scan records whether the prefix is a valid JSON prefix, its nesting depth, whether it ends inside a string and under which tool-argument key, and, within string contents, unmatched brackets, quote parity, open code fences, heredocs, and the termination pattern of the last line.

\noindent\textbf{Prefix recall.}
Earlier completed model requests from the same run serve as references, each represented by its hashed prefix at every checkpoint and its final generated length. The current prefix is compared with these references using cosine similarity, and the record includes the final and remaining lengths associated with the most similar requests and checkpoints.

\noindent\textbf{Stream timing.}
For the current request, let $s$ denote its start time and let $c_1<\cdots<c_m$ denote the output-checkpoint times. The timing record contains the time to the first checkpoint $c_1-s$, elapsed time $c_m-s$, streaming time $c_m-c_1$, committed bytes divided by streaming time, time since the previous checkpoint, the longest gap between checkpoints, and the initial-delay fraction $(c_1-s)/(c_m-s)$. Fields with undefined denominators or too few observations are masked.

\FloatBarrier
\subsection{Predictors and Training}
\label{app:segment_predictor}

\noindent\textbf{Predictors.}
TokenCast uses separate LightGBM models for the four prediction settings. Call Start and In-call Update predict the complete consumption of the current call. Task Start and Task Update use a direct path and a compositional path. The direct path predicts remaining task consumption as one target. The compositional path predicts the current or next segment, its ending boundary, and the subsequent suffix, then combines the two segment representations using Eq.~(\ref{eq:segment_composition}).

\noindent\textbf{Training instances.}
Labels are extracted from completed traces. Call-level instances use the realized complete-call consumption. Task-level instances contain the remaining-consumption target, the prefix representation and ending boundary, and the suffix representation. Model inputs contain only the evidence available at the corresponding prediction point.

\noindent\textbf{Task-level fitting.}
The direct and prefix models are fitted first. Task-level cross-fitting then produces out-of-fold prefix boundaries for suffix training. The suffix residual label is recomputed against the predicted input baseline as described in Section~\ref{sec:compositional_learning}. After these models are fixed, the complete pipeline produces out-of-fold forecasts used to fit the correction model in Eq.~(\ref{eq:task_correction}).

\noindent\textbf{Inference.}
Call-level forecasts use the corresponding fitted model directly. Task-level forecasts compute the direct prediction, the prefix boundary, the suffix prediction, and the compositional forecast before applying the correction model. The point models remain fixed during execution. Raw interval endpoints are produced by the quantile models in Appendix~\ref{app:interval_calibration}.

\FloatBarrier
\subsection{Theoretical Analysis}
\label{app:theoretical_analysis}

\noindent\textbf{Composition properties.}
Let $L_A$ and $L_B$ be the initial input lengths of adjacent segments $A$ and $B$. Since $L_B=L_A+g_A$ and $C_X=n_XL_X+b_X$ for $X\in\{A,B\}$,
\[
C_{A\circ B}
=(n_A+n_B)L_A+b_A+b_B+n_Bg_A
=C_A+C_B.
\]
The cross-term $n_Bg_A$ arises when the cost of $B$ is expressed relative to the initial input length of $A$. For three consecutive segments $A$, $B$, and $D$, either grouping yields the residual $b_A+b_B+b_D+n_Bg_A+n_D(g_A+g_B)$. The composition is therefore associative.

\noindent\textbf{Single-call representation.}
A single call has representation $\phi_j=(1,\,L_{j+1}-L_j,\,C_j-L_j)$. For the terminal call, set $L_{K+1}=L_K$ to close the notation; no suffix follows it. Composing these representations in execution order recovers the representation of any contiguous segment. For a complete trace of $K$ calls, the resulting residual is $\sum_{j=1}^{K}C_j-KL_1$. Hence
\[
T=KL_1+b_{\phi_1\circ\cdots\circ\phi_K},
\]
regardless of how the trace is partitioned.

\noindent\textbf{Error propagation and loss weights.}
In a compositional forecast, the suffix is predicted from the estimated boundary of the prefix. Let $L_B=L_A+g_A$ be the true input length at the start of the suffix, and let $\delta g_A$ and $\delta n_B$ denote prediction errors in the prefix's context change and the suffix's call count. With $L_A$ fixed, and writing $\delta b_B$ for the error in the suffix residual, the suffix cost error is
\[
\Delta C_B
=n_B\delta g_A+L_B\delta n_B
+\delta n_B\delta g_A+\delta b_B.
\]
The coefficients $n_B$ and $L_B$ motivate the local weights in Eq.~(\ref{eq:local_cost_weighting}) under a true-boundary residual. Suffix training instead uses $b_B^{\mathrm{train}}=C_B-n_B\widehat{L}_B$. Writing $e_b=\widehat b_B-b_B^{\mathrm{train}}$ and $e_n=\widehat n_B-n_B$ gives $\widehat C_B-C_B=\widehat L_B e_n+e_b$. Boundary prediction can affect the learned residual and the suffix model's inputs. Cross-fitting exposes suffix training to out-of-fold boundary errors.

\FloatBarrier
\subsection{Interval Calibration}
\label{app:interval_calibration}

After the point models are fixed, separate LightGBM quantile models are trained for the 0.05 and 0.95 endpoints at each prediction setting. They use the evidence available at that prediction point and are fitted on the training tasks. Their outputs form the raw interval $[\ell_i,u_i]$.

Intervals are calibrated separately for Task Start, Call Start, In-call Update, and Task Update using the held-out calibration tasks of each fold. For target $y_i$, the absolute interval residual is
\begin{equation}
    s_i=\max\left\{\ell_i-y_i,\;y_i-u_i,\;0\right\}.
\end{equation}
For each prediction setting, the corresponding calibration quantile of $\{s_i\}$ is added symmetrically to the raw endpoints. At runtime, the calibrated endpoints are bounded below by consumption already confirmed within the prediction scope. Calibration parameters remain fixed during inference.

\FloatBarrier
\section{Experimental Setup}
\label{app:experimental_details}

\FloatBarrier
\subsection{Trace Collection}
\label{app:collection}

\noindent\textbf{Benchmarks.}
From the 500 instances of SWE-bench Verified~\citep{DBLP:conf/iclr/JimenezYWYPPN24, chowdhury2024swebenchverified}, we take 144. The four smallest repositories, flask, seaborn, requests, and pylint, enter in full. pytest, xarray, astropy, scikit-learn, and matplotlib contribute 15 each, and sphinx, sympy, and django 16 each. Search-R1~\citep{DBLP:journals/corr/abs-2503-09516} contributes 32 questions from seven FlashRAG evaluation sets~\citep{DBLP:conf/www/Jin0DDYZZYW25}, half single-hop from NQ~\citep{kwiatkowski2019natural}, TriviaQA~\citep{joshi2017triviaqa}, and PopQA~\citep{mallen2023not}, and half multi-hop from HotpotQA~\citep{yang2018hotpotqa}, 2WikiMultihopQA~\citep{ho2020constructing}, MuSiQue~\citep{trivedi2022musique}, and Bamboogle~\citep{press2023measuring}. MMLU-Pro~\citep{DBLP:conf/nips/WangMZNCGRAHJLK24} contributes 32 test questions with two or three per category, and LongBench-v2~\citep{DBLP:conf/acl/BaiTZ0WLCX0D0L25} has 32 questions spread over its six domains and three length labels. Within each stratum, we take instances in the stable-hash order of their id, and the anchors are the first two per SWE-bench Verified repository, one for flask and three for django, and the first eight in each other benchmark.

\noindent\textbf{Task packs.}
A task is a statement and a seed workspace. For SWE-bench, the seed is the repository at the base commit. The agent may not edit tests and finishes by writing \texttt{submission.patch}, the diff of its working tree. For the other benchmarks, the seed holds \texttt{README.md} and an empty \texttt{answer.md}, the statement provides the question and its options, for LongBench-v2 together with the context document, and the agent writes its answer into \texttt{answer.md}. A Search-R1 statement also names a shell command that queries a local BM25 server~\citep{robertson2009probabilistic, DBLP:journals/corr/abs-2503-09516}, the official Search-R1 retriever over the 21,015,324-passage wiki-18 corpus, and prints three passages. Correctness is scored after collection, with the SWE-bench Verified harness for patches, letter match for MMLU-Pro and LongBench-v2, and alias match for Search-R1.

\noindent\textbf{Agent LLMs.}
Table~\ref{tab:agent_llms} lists the models and their reasoning configurations. GPT-5.4~\citep{openai2026gpt54}, Claude Opus 4.6~\citep{anthropic2026claudeopus46}, and Gemini 3.1 Pro~\citep{googledeepmind2026gemini31pro} use the lowest reasoning setting available through their respective APIs. DeepSeek-V4-Pro~\citep{deepseekai2026deepseekv4} and Qwen3.8-27B~\citep{qwen38} run with thinking enabled, while Llama-3.2-3B-Instruct~\citep{meta2024llama32} has no reasoning mode. Temperature is 0 where supported. The self-hosted models run on vLLM~\citep{DBLP:conf/sosp/KwonLZ0ZY0ZS23} in bf16 on eight A100 GPUs with full context windows.

\begin{table}[htbp]
\centering
\small
\caption{Agent LLMs used for trace collection, with access modes and reasoning configurations.}
\label{tab:agent_llms}
\begin{tabular}{llll}
\toprule
Model & Vendor & Access & Reasoning configuration \\
\midrule
GPT-5.4 & OpenAI & API & Low reasoning effort \\
Claude Opus 4.6 & Anthropic & API & Adaptive thinking; low effort \\
Gemini 3.1 Pro & Google & API & Low thinking level \\
DeepSeek-V4-Pro & DeepSeek & API & Thinking enabled \\
Qwen3.8-27B & Alibaba & Self-hosted & Thinking enabled \\
Llama-3.2-3B-Instruct & Meta & Self-hosted & No reasoning mode \\
\bottomrule
\end{tabular}
\end{table}

\noindent\textbf{Harnesses.}
DeepSeek Harness~\citep{deepseekharness} runs at version 0.1.1-rc.2, commit \texttt{b150a551}, in its headless profile with 12 tools: \texttt{read}, \texttt{write}, \texttt{edit}, \texttt{str\_replace\_editor}, \texttt{glob}, \texttt{grep}, \texttt{pwsh}, \texttt{job\_list}, \texttt{job\_output}, \texttt{job\_kill}, \texttt{todo\_write}, and \texttt{skill}. Commands are executed without a sandbox or approval prompts. Context compaction, result pruning, generated session titles, web access, and subagents are turned off, so every provider request is an agent call. Each run starts a fresh session in a fresh copy of the seed workspace on Windows.

OpenHands~\citep{DBLP:conf/iclr/0001LSXTZPSLSTL25} runs at version 1.18.0 with the CodeAct~\citep{wang2024executable} agent and its \texttt{execute\_bash}, \texttt{str\_replace\_editor}, and task-tracking tools, with browsing and the condenser disabled. It runs SWE-bench Verified in the official image of each instance and the other benchmarks in a plain Linux image and calls the same retrieval script.

Both harnesses cap a run at 7,200 s, 500 calls, 4 retries per call, 65,536 output tokens per model request, and 30,000,000 tokens. A run ends when the agent submits, the model gives a final response without submitting, or a cap is reached. A logical call consists of its initial model request and any retry requests issued before tool feedback is received. The observer timestamps every model request and emits a checkpoint every 128 bytes of committed output. Token usage is taken from the provider response for each request and aggregated at the logical-call level.

\noindent\textbf{Repeated executions.}
Table~\ref{tab:trace_design} lists the runs per task. SWE-bench Verified tasks run twice per model, and the other tasks run four times. Anchors run eight times, or 16 for Search-R1.

\noindent\textbf{Trace statistics.}
Table~\ref{tab:trace_stats} lists the median tokens and calls per run and the share of correct runs for each benchmark and model. For GPT-5.4 on SWE-bench Verified, the input makes up 99\% of the tokens, the median run takes 129 s, and the median call emits two checkpoints.

\begin{table}[htbp]
\centering
\small
\caption{Median tokens in thousands and calls among runs with complete usage, and correct runs as a percentage of all attempted runs, under each harness.}
\label{tab:trace_stats}
\begin{tabular}{lcccccc}
\toprule
& \multicolumn{3}{c}{DeepSeek Harness} & \multicolumn{3}{c}{OpenHands} \\
\cmidrule(lr){2-4}\cmidrule(lr){5-7}
Model & Tokens & Calls & Correct & Tokens & Calls & Correct \\
\midrule
\multicolumn{7}{l}{\textit{SWE-bench Verified}} \\
GPT-5.4 & 245.4 & 21 & 40.7 & 314.5 & 20 & 36.3 \\
Claude Opus 4.6 & 258.1 & 21 & 41.7 & 470.7 & 22 & 54.2 \\
Gemini 3.1 Pro & 333.5 & 21 & 43.5 & 253.5 & 21 & 47.5 \\
DeepSeek-V4-Pro & 280.9 & 20 & 34.3 & 336.8 & 28 & 33.8 \\
Qwen3.8-27B & 526.0 & 22 & 29.6 & 211.5 & 15 & 31.2 \\
Llama-3.2-3B-Instruct & 508.8 & 31 & 9.5 & 619.7 & 33 & 11.8 \\
\midrule
\multicolumn{7}{l}{\textit{Search-R1}} \\
GPT-5.4 & 38.1 & 8 & 76.3 & 37.5 & 8 & 75.0 \\
Claude Opus 4.6 & 58.3 & 9 & 59.8 & 58.5 & 10 & 69.6 \\
Gemini 3.1 Pro & 55.8 & 10 & 68.8 & 48.3 & 8 & 69.6 \\
DeepSeek-V4-Pro & 94.8 & 12 & 65.2 & 41.2 & 9 & 70.5 \\
Qwen3.8-27B & 60.3 & 10 & 54.0 & 84.0 & 9 & 58.9 \\
Llama-3.2-3B-Instruct & 136.4 & 16 & 27.2 & 98.7 & 15 & 28.6 \\
\midrule
\multicolumn{7}{l}{\textit{MMLU-Pro}} \\
GPT-5.4 & 21.6 & 4 & 81.9 & 46.1 & 6 & 81.9 \\
Claude Opus 4.6 & 27.3 & 5 & 87.5 & 29.3 & 4 & 88.1 \\
Gemini 3.1 Pro & 54.6 & 9 & 75.6 & 59.7 & 9 & 77.5 \\
DeepSeek-V4-Pro & 35.2 & 6 & 70.6 & 38.6 & 5 & 76.9 \\
Qwen3.8-27B & 41.9 & 7 & 76.9 & 52.7 & 7 & 67.5 \\
Llama-3.2-3B-Instruct & 30.1 & 5 & 42.5 & 48.7 & 5 & 43.8 \\
\midrule
\multicolumn{7}{l}{\textit{LongBench-v2}} \\
GPT-5.4 & 762.5 & 6 & 58.8 & 608.0 & 5 & 58.1 \\
Claude Opus 4.6 & 361.7 & 4 & 66.9 & 716.1 & 6 & 75.0 \\
Gemini 3.1 Pro & 832.5 & 8 & 64.4 & 610.0 & 8 & 69.4 \\
DeepSeek-V4-Pro & 523.7 & 4 & 60.6 & 1,030.7 & 9 & 52.5 \\
Qwen3.8-27B & 884.4 & 8 & 45.0 & 581.0 & 6 & 45.6 \\
Llama-3.2-3B-Instruct & 645.9 & 7 & 25.6 & 592.1 & 4 & 13.1 \\
\bottomrule
\end{tabular}
\end{table}

The code-repair execution in Appendix~\ref{app:case_code} illustrates the four prediction points and the usage records along a complete run (Figure~\ref{fig:example_run}).

\noindent\textbf{Additional trajectories.}
The generalization experiments use independently released trajectories from LiveClawBench~\citep{long2026liveclawbench}, which records multiple agent LLMs across task domains under a shared agent framework. We retain runs with complete interaction records and per-call input and output token counts, and construct targets with the accounting convention of Section~\ref{sec:problem_formulation}.

For the reasoning-configuration experiment in Appendix~\ref{app:reasoning_off}, we rerun 69 SWE-bench Verified tasks with GPT-5.4 reasoning effort set to off and compare them with the same tasks under the low setting. The API route reports no separate reasoning-token count. Measured against visible output, the median request bills 0.46 output tokens per byte under the low setting and 0.34 under off.

\FloatBarrier
\subsection{Data Splits}
\label{app:data_splits}

Tasks are assigned to five folds, balanced over benchmarks and agent LLMs, and all runs of a task remain in their fold. Each fold serves once as the test fold. The next fold calibrates intervals, the one after selects hyperparameters, and the remaining two train. We repeat the partitioning with three random seeds. For each seed, we pool out-of-fold test predictions and compute task-weighted metrics over all evaluated tasks. Tables~\ref{tab:main} and~\ref{tab:full_swebench}--\ref{tab:full_longbench} report means across the three seeds.

\FloatBarrier
\subsection{Baselines}
\label{app:baselines}

\noindent\textbf{Output-length predictors.}
TRAIL, EGTP, and TIE are originally designed to predict the length of a single response. We adapt each method to the target associated with a prediction point while preserving its core representation and estimator. For Qwen3.8-27B and Llama-3.2-3B-Instruct, the internal-state methods read the last layer of the agent LLM itself. For the API models, they use Llama-3.2-3B-Instruct as a proxy encoder over the text visible at the prediction point: the task statement at Task Start, the statement and transcript of completed calls at Call Start and Task Update, and the statement and generated prefix at In-call Update. The window keeps the first 384 tokens of the statement and the last 1,536 of the remaining text. TRAIL classifies the last-token state into 8 log-spaced target-length bins and predicts the expected bin median. EGTP fits a ridge regression of log target length on the entropy-weighted mean state and entropy statistics of the window. TIE encodes the window with bge-small-en-v1.5~\citep{DBLP:conf/sigir/XiaoLZMLN24} and fits the location and scale of a log-$t$ distribution with 3.5 degrees of freedom on the concatenated CLS, mean, and max poolings. Its point prediction is the median, and its interval spans the 0.05 and 0.95 quantiles.

\noindent\textbf{Agent usage estimators.}
Self-Prediction queries the agent LLM for the 5th, 50th, and 95th percentiles of token consumption for the same prediction target. The median serves as the point prediction, and the 5th and 95th percentiles define a nominal 90\% prediction interval. We use these predictions directly, without fitting or calibrating them on held-out tasks. At Task Start, the agent first inspects the seed workspace with its full tool set and estimates the tokens of a complete run. At the other prediction points, we adapt the prompt to the observed execution record and corresponding prediction target. At Call Start, the agent estimates the complete consumption of the current call. At In-call Update, it estimates the same complete-call quantity using the observed generation prefix. At Task Update, it estimates the token consumption of the remaining execution. Appendix~\ref{app:self_prediction_prompt} provides the complete prompt template for all four points.

\noindent\textbf{Fitting.}
Each learned baseline uses TokenCast's partition: it is fitted on each fold's training tasks, tuned on the validation tasks, and calibrated on the calibration tasks.

\FloatBarrier
\subsection{Metrics}
\label{app:metrics}

For each prediction setting, let $i$ index the evaluated prediction points, with target $y_i$, point prediction $\hat{y}_i$, 90\% prediction interval $[\ell_i,u_i]$, and normalized weight $w_i$. We assign equal weight to each evaluated task, divide that weight equally among its evaluated runs, and divide each run's weight equally among its evaluated prediction points. Thus, $\sum_i w_i=1$, and
\begin{equation}
\mathrm{MAE}=\sum_i w_i|\hat{y}_i-y_i|,
\qquad
\bar{y}=\sum_i w_i y_i,
\qquad
\mathrm{WAPE}=\frac{\mathrm{MAE}}{\bar{y}}\times100\%.
\end{equation}
MAE and mean target consumption $\bar{y}$ use the same evaluation samples and task--run--checkpoint weights. WAPE expresses the mean absolute error as a percentage of the mean actual target consumption.
For a nominal $1-\alpha$ prediction interval, we use the interval score
\begin{equation}
\mathrm{IS}_i=(u_i-\ell_i)+\frac{2}{\alpha}(\ell_i-y_i)\mathbf{1}\{y_i<\ell_i\}+\frac{2}{\alpha}(y_i-u_i)\mathbf{1}\{y_i>u_i\},
\end{equation}
with $\alpha=0.1$. Mean interval score is
\begin{equation}
\mathrm{MIS}=\sum_i w_i\,\mathrm{IS}_i.
\end{equation}
Empirical coverage and mean interval width are aggregated using the same weights. The Norm.\ Avg.\ column of Table~\ref{tab:main} is the arithmetic mean over the four prediction points of a method's MAE divided by the MAE of the corresponding history-median predictor. This predictor outputs the median target of the training tasks in the same fold for the same benchmark, agent LLM, and prediction point, and uses no task features or execution evidence.

\FloatBarrier
\section{Extended Evaluation}
\label{app:extended_evaluation}

\FloatBarrier
\subsection{Run-to-Run Variation and Interval Reliability}
\label{app:prediction_reliability}

\begin{figure}[htbp]
\centering
\begin{subfigure}[t]{0.412\textwidth}
    \centering
    \includegraphics[width=\linewidth]{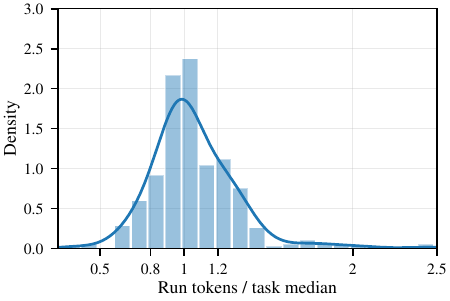}
    \caption{Run-to-run variation}
\end{subfigure}
\hfill
\begin{subfigure}[t]{0.5\textwidth}
    \centering
    \includegraphics[width=\linewidth]{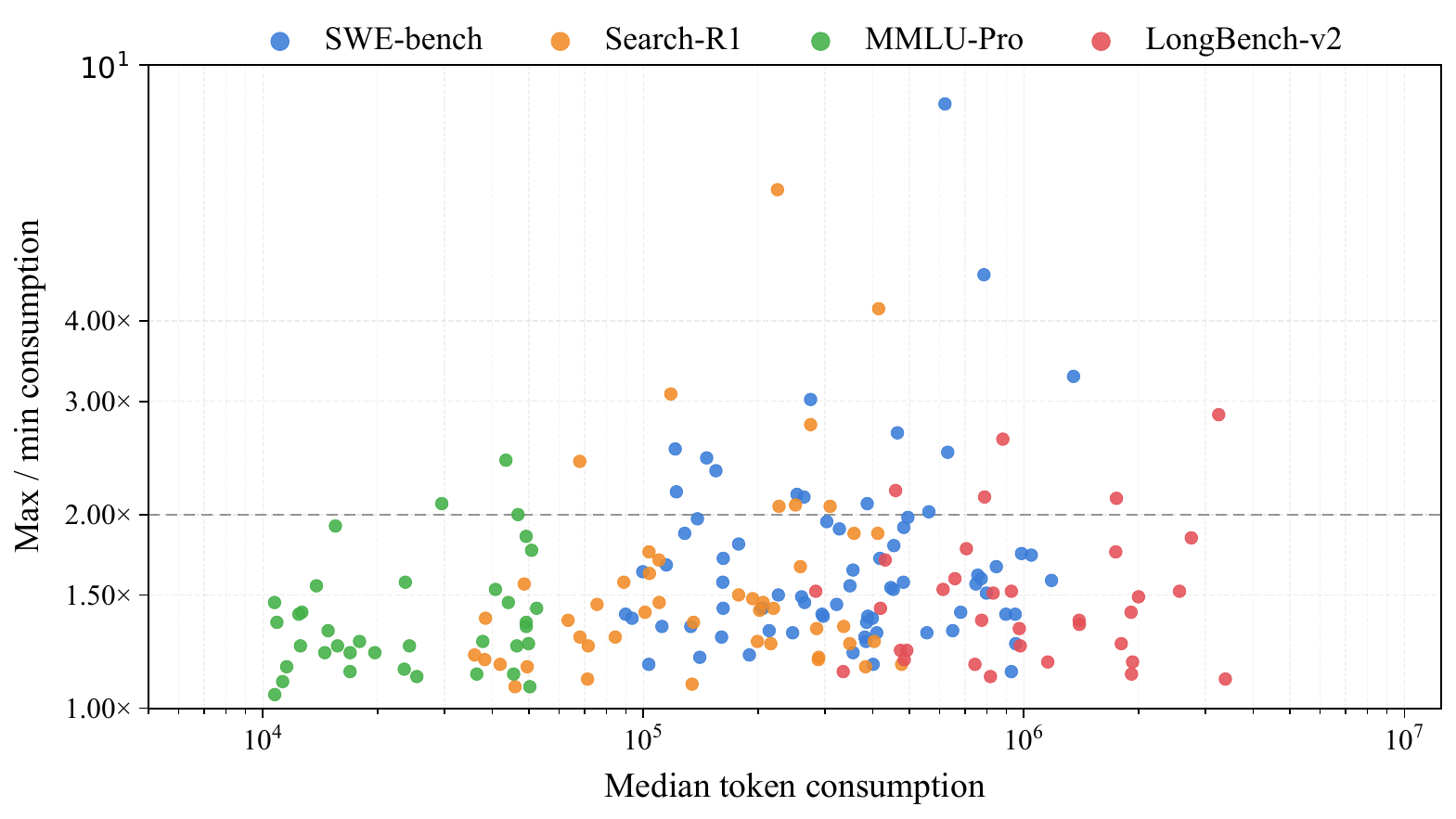}
    \caption{Consumption spread}
\end{subfigure}
\caption{Run-to-run variation in token consumption across repeated executions of the same task. The panels show the distribution across repeated runs and the within-task consumption spread as a function of task consumption, over 48 GPT-5.4 anchor tasks.}
\label{fig:repeated_runs}
\end{figure}

Figure~\ref{fig:repeated_runs} summarizes repeated GPT-5.4 executions from 48 anchor tasks across four benchmarks and two harnesses. Search-R1 has 16 repetitions per task--harness pair, while the others have eight. Repeated executions of a task show wide consumption spread across all four benchmarks.

Table~\ref{tab:repeated_runs} compares TokenCast's calibrated intervals with the native intervals returned by Self-Prediction at the four prediction points. MIS combines interval width with a penalty for how far outcomes fall outside the interval. Lower values indicate better interval forecasts. TokenCast's wider intervals reduce the missed-outcome penalty enough to improve MIS. This comparison includes TokenCast's held-out calibration and Self-Prediction's native, uncalibrated intervals.

\FloatBarrier
\subsection{Results Across Benchmarks and Agent LLMs}
\label{app:benchmark_results}

Tables~\ref{tab:full_swebench} to~\ref{tab:full_longbench} report the MAE for each benchmark and agent LLM on the same runs, with the history median as the normalization reference. MAE is reported in tokens, with k denoting thousands. Norm.\ Avg.\ follows the normalization in Appendix~\ref{app:metrics}.

\begin{table}[htbp]
\centering
\small
\caption{MAE on SWE-bench Verified. Norm.\ Avg.\ is the macro-average of MAE normalized at each prediction point by the history-median MAE. $k$ denotes thousands of tokens. Best and second-best results are in bold and underlined, respectively. Marks are assigned using unrounded values.}
\label{tab:full_swebench}
\setlength{\tabcolsep}{5pt}
\begin{tabular}{lrrrrrr}
\toprule
Prediction point & TokenCast & TRAIL & EGTP & TIE & Self-Pred. & History median \\
\midrule
\multicolumn{7}{l}{\textit{GPT-5.4}} \\
Task Start & \textbf{144.0k} & 165.3k & 160.9k & 157.0k & \underline{152.0k} & 179.0k \\
Call Start & \textbf{64.5} & 71.1 & 70.3 & \underline{69.5} & 70.9 & 74.2 \\
In-call Update & \textbf{38.9} & 78.9 & \underline{74.6} & 77.4 & 80.2 & 80.3 \\
Task Update & \textbf{80.0k} & \underline{115.0k} & 117.0k & 119.0k & 126.0k & 130.0k \\
\rowcolor{gray!12}
Norm. Avg. & \textbf{0.69} & 0.94 & \underline{0.92} & 0.92 & 0.94 & 1.00 \\
\midrule
\multicolumn{7}{l}{\textit{Claude Opus 4.6}} \\
Task Start & 150.7k & 157.5k & 153.8k & \underline{146.8k} & \textbf{138.9k} & 171.0k \\
Call Start & \textbf{64.1} & 69.4 & 69.9 & \underline{66.8} & 68.8 & 73.5 \\
In-call Update & \textbf{55.0} & 80.2 & \underline{71.7} & 74.4 & 78.1 & 83.4 \\
Task Update & \textbf{88.7k} & \underline{112.7k} & 122.4k & 116.4k & 122.0k & 133.0k \\
\rowcolor{gray!12}
Norm. Avg. & \textbf{0.77} & 0.92 & 0.91 & \underline{0.88} & 0.90 & 1.00 \\
\midrule
\multicolumn{7}{l}{\textit{Gemini 3.1 Pro}} \\
Task Start & \textbf{157.8k} & 173.5k & 170.2k & 176.4k & \underline{163.6k} & 193.0k \\
Call Start & \textbf{68.3} & 74.7 & 71.6 & \underline{68.4} & 79.8 & 88.5 \\
In-call Update & \textbf{41.0} & 84.3 & 79.4 & \underline{77.2} & 87.2 & 96.2 \\
Task Update & \textbf{69.6k} & 126.3k & 121.4k & \underline{118.6k} & 132.2k & 151.0k \\
\rowcolor{gray!12}
Norm. Avg. & \textbf{0.62} & 0.86 & 0.83 & \underline{0.82} & 0.88 & 1.00 \\
\midrule
\multicolumn{7}{l}{\textit{DeepSeek-V4-Pro}} \\
Task Start & 212.9k & 204.9k & 211.5k & \underline{198.4k} & \textbf{194.5k} & 232.0k \\
Call Start & \textbf{80.0} & 89.0 & \underline{84.6} & 85.9 & 91.3 & 101.5 \\
In-call Update & \textbf{65.5} & 96.1 & 94.9 & \underline{86.3} & 104.8 & 112.8 \\
Task Update & \textbf{106.2k} & \underline{143.4k} & 155.6k & 148.1k & 158.4k & 177.0k \\
\rowcolor{gray!12}
Norm. Avg. & \textbf{0.72} & 0.86 & 0.87 & \underline{0.83} & 0.89 & 1.00 \\
\midrule
\multicolumn{7}{l}{\textit{Qwen3.8-27B}} \\
Task Start & \textbf{192.0k} & 232.0k & \underline{209.0k} & 215.0k & 221.0k & 284.0k \\
Call Start & \textbf{87.0} & 109.0 & 102.9 & \underline{99.0} & 105.5 & 119.0 \\
In-call Update & \textbf{81.9} & 120.0 & 116.0 & 111.4 & \underline{108.9} & 147.0 \\
Task Update & \textbf{124.0k} & 176.0k & 171.0k & \underline{164.0k} & 173.0k & 198.0k \\
\rowcolor{gray!12}
Norm. Avg. & \textbf{0.65} & 0.86 & 0.81 & \underline{0.79} & 0.82 & 1.00 \\
\midrule
\multicolumn{7}{l}{\textit{Llama-3.2-3B-Instruct}} \\
Task Start & 272.3k & 277.6k & 269.7k & \textbf{258.2k} & \underline{264.8k} & 302.0k \\
Call Start & \textbf{133.8} & 140.7 & \underline{134.6} & 140.0 & 141.6 & 147.0 \\
In-call Update & \textbf{100.3} & 141.8 & 137.7 & \underline{126.4} & 150.2 & 158.0 \\
Task Update & \textbf{152.2k} & 205.3k & 213.1k & \underline{188.2k} & 213.7k & 218.0k \\
\rowcolor{gray!12}
Norm. Avg. & \textbf{0.79} & 0.93 & 0.91 & \underline{0.87} & 0.94 & 1.00 \\
\bottomrule
\end{tabular}
\end{table}

\begin{table}[htbp]
\centering
\small
\caption{MAE on Search-R1. Norm.\ Avg.\ is the macro-average of MAE normalized at each prediction point by the history-median MAE. $k$ denotes thousands of tokens. Best and second-best results are in bold and underlined, respectively. Marks are assigned using unrounded values.}
\label{tab:full_searchr1}
\setlength{\tabcolsep}{5pt}
\begin{tabular}{lrrrrrr}
\toprule
Prediction point & TokenCast & TRAIL & EGTP & TIE & Self-Pred. & History median \\
\midrule
\multicolumn{7}{l}{\textit{GPT-5.4}} \\
Task Start & \textbf{34.2k} & 37.8k & 37.6k & \underline{36.4k} & 36.9k & 44.9k \\
Call Start & \underline{34.5} & 37.3 & 36.5 & \textbf{34.0} & 39.8 & 44.0 \\
In-call Update & \textbf{31.3} & 39.4 & 38.5 & \underline{36.8} & 39.9 & 42.0 \\
Task Update & \textbf{17.6k} & \underline{23.1k} & 23.7k & 23.3k & 24.2k & 25.0k \\
\rowcolor{gray!12}
Norm. Avg. & \textbf{0.75} & 0.89 & 0.88 & \underline{0.85} & 0.91 & 1.00 \\
\midrule
\multicolumn{7}{l}{\textit{Claude Opus 4.6}} \\
Task Start & \underline{31.1k} & 33.9k & 32.9k & 32.2k & \textbf{29.8k} & 39.8k \\
Call Start & \textbf{30.4} & 33.6 & 32.0 & \underline{30.4} & 34.9 & 40.2 \\
In-call Update & \textbf{22.8} & 33.4 & 31.6 & \underline{29.7} & 34.0 & 38.5 \\
Task Update & \textbf{12.9k} & 20.8k & \underline{18.7k} & 19.5k & 21.4k & 23.6k \\
\rowcolor{gray!12}
Norm. Avg. & \textbf{0.67} & 0.86 & 0.81 & \underline{0.79} & 0.85 & 1.00 \\
\midrule
\multicolumn{7}{l}{\textit{Gemini 3.1 Pro}} \\
Task Start & \underline{38.0k} & 42.3k & 41.0k & 40.7k & \textbf{37.1k} & 49.8k \\
Call Start & \underline{40.0} & 44.7 & \textbf{39.8} & 42.3 & 47.0 & 54.5 \\
In-call Update & \textbf{25.5} & 46.8 & 44.2 & \underline{42.6} & 47.5 & 57.0 \\
Task Update & \textbf{16.1k} & 27.8k & \underline{24.6k} & 25.2k & 28.6k & 31.5k \\
\rowcolor{gray!12}
Norm. Avg. & \textbf{0.61} & 0.84 & \underline{0.78} & 0.79 & 0.84 & 1.00 \\
\midrule
\multicolumn{7}{l}{\textit{DeepSeek-V4-Pro}} \\
Task Start & 46.5k & 50.0k & 47.5k & \underline{44.2k} & \textbf{42.9k} & 58.5k \\
Call Start & \underline{50.7} & 57.3 & 54.0 & \textbf{50.6} & 60.6 & 68.0 \\
In-call Update & \textbf{35.1} & 60.8 & 56.2 & \underline{50.4} & 62.3 & 71.5 \\
Task Update & \textbf{19.9k} & 33.7k & 31.1k & \underline{29.8k} & 34.5k & 37.2k \\
\rowcolor{gray!12}
Norm. Avg. & \textbf{0.64} & 0.86 & 0.81 & \underline{0.75} & 0.86 & 1.00 \\
\midrule
\multicolumn{7}{l}{\textit{Qwen3.8-27B}} \\
Task Start & \textbf{39.0k} & 53.5k & 49.8k & \underline{47.6k} & 51.0k & 72.9k \\
Call Start & \textbf{56.7} & \underline{65.9} & 71.4 & 67.7 & 76.1 & 91.0 \\
In-call Update & \textbf{57.0} & 84.6 & 80.7 & \underline{75.5} & 83.2 & 93.0 \\
Task Update & \textbf{24.0k} & 39.5k & \underline{34.0k} & 35.4k & 38.4k & 41.0k \\
\rowcolor{gray!12}
Norm. Avg. & \textbf{0.59} & 0.83 & 0.79 & \underline{0.77} & 0.84 & 1.00 \\
\midrule
\multicolumn{7}{l}{\textit{Llama-3.2-3B-Instruct}} \\
Task Start & 59.0k & 60.0k & 57.9k & \underline{53.6k} & \textbf{52.4k} & 68.0k \\
Call Start & 79.0 & 80.9 & \textbf{70.4} & \underline{75.4} & 84.6 & 95.0 \\
In-call Update & \textbf{64.0} & 99.4 & 91.2 & \underline{84.7} & 95.6 & 108.0 \\
Task Update & \textbf{31.0k} & 43.8k & 40.2k & \underline{38.6k} & 45.9k & 49.0k \\
\rowcolor{gray!12}
Norm. Avg. & \textbf{0.73} & 0.89 & 0.81 & \underline{0.79} & 0.87 & 1.00 \\
\bottomrule
\end{tabular}
\end{table}

\begin{table}[htbp]
\centering
\small
\caption{MAE on MMLU-Pro. Norm.\ Avg.\ is the macro-average of MAE normalized at each prediction point by the history-median MAE. $k$ denotes thousands of tokens. Best and second-best results are in bold and underlined, respectively. Marks are assigned using unrounded values.}
\label{tab:full_mmlupro}
\setlength{\tabcolsep}{5pt}
\begin{tabular}{lrrrrrr}
\toprule
Prediction point & TokenCast & TRAIL & EGTP & TIE & Self-Pred. & History median \\
\midrule
\multicolumn{7}{l}{\textit{GPT-5.4}} \\
Task Start & \underline{8.0k} & 8.8k & 8.5k & 8.2k & \textbf{7.4k} & 9.7k \\
Call Start & \textbf{14.8} & 16.7 & \underline{15.5} & 15.8 & 17.6 & 19.6 \\
In-call Update & \textbf{9.2} & 19.6 & 18.3 & \underline{16.7} & 17.4 & 24.3 \\
Task Update & \textbf{4.3k} & 6.2k & 6.5k & 6.0k & \underline{5.8k} & 6.9k \\
\rowcolor{gray!12}
Norm. Avg. & \textbf{0.65} & 0.87 & 0.84 & \underline{0.80} & 0.80 & 1.00 \\
\midrule
\multicolumn{7}{l}{\textit{Claude Opus 4.6}} \\
Task Start & 7.5k & 8.0k & 7.7k & \underline{7.3k} & \textbf{6.9k} & 8.9k \\
Call Start & 14.9 & \underline{14.8} & 15.6 & \textbf{14.3} & 16.1 & 18.3 \\
In-call Update & \textbf{11.5} & \underline{15.5} & 16.8 & 16.0 & 18.4 & 22.0 \\
Task Update & \textbf{4.4k} & 6.0k & 6.1k & 5.8k & \underline{5.6k} & 6.3k \\
\rowcolor{gray!12}
Norm. Avg. & \textbf{0.72} & 0.84 & 0.86 & \underline{0.81} & 0.85 & 1.00 \\
\midrule
\multicolumn{7}{l}{\textit{Gemini 3.1 Pro}} \\
Task Start & \textbf{8.6k} & 10.1k & 9.6k & 9.3k & \underline{8.9k} & 11.5k \\
Call Start & \textbf{18.0} & \underline{18.3} & 20.1 & 19.4 & 21.3 & 23.9 \\
In-call Update & \textbf{10.5} & 24.7 & \underline{20.6} & 21.8 & 23.6 & 28.7 \\
Task Update & \textbf{4.7k} & 7.2k & 6.8k & \underline{6.4k} & 7.5k & 8.0k \\
\rowcolor{gray!12}
Norm. Avg. & \textbf{0.61} & 0.85 & 0.81 & \underline{0.79} & 0.86 & 1.00 \\
\midrule
\multicolumn{7}{l}{\textit{DeepSeek-V4-Pro}} \\
Task Start & 11.6k & 12.3k & 11.9k & \underline{11.0k} & \textbf{10.6k} & 13.7k \\
Call Start & \underline{24.4} & 26.2 & 24.8 & \textbf{23.5} & 26.9 & 29.6 \\
In-call Update & \textbf{19.5} & 31.8 & 28.7 & \underline{25.4} & 30.4 & 35.9 \\
Task Update & \textbf{6.9k} & \underline{8.5k} & 9.2k & 8.9k & 9.4k & 9.8k \\
\rowcolor{gray!12}
Norm. Avg. & \textbf{0.73} & 0.88 & 0.86 & \underline{0.80} & 0.87 & 1.00 \\
\midrule
\multicolumn{7}{l}{\textit{Qwen3.8-27B}} \\
Task Start & \underline{13.0k} & 14.3k & \textbf{12.5k} & 13.7k & 13.3k & 16.2k \\
Call Start & \textbf{26.5} & 30.3 & 28.6 & \underline{27.9} & 31.1 & 35.1 \\
In-call Update & \textbf{21.1} & 37.7 & 35.5 & 34.3 & \underline{32.8} & 42.4 \\
Task Update & \textbf{8.4k} & 11.2k & \underline{10.3k} & 10.7k & 11.5k & 12.0k \\
\rowcolor{gray!12}
Norm. Avg. & \textbf{0.69} & 0.89 & \underline{0.82} & 0.84 & 0.86 & 1.00 \\
\midrule
\multicolumn{7}{l}{\textit{Llama-3.2-3B-Instruct}} \\
Task Start & 16.4k & 17.5k & 16.9k & \textbf{15.4k} & \underline{15.8k} & 19.6k \\
Call Start & \underline{33.5} & 36.5 & \textbf{33.2} & 33.8 & 37.9 & 41.8 \\
In-call Update & \textbf{30.7} & 46.0 & 43.7 & \underline{42.3} & 47.3 & 50.2 \\
Task Update & \textbf{10.8k} & 13.2k & 13.8k & 13.5k & \underline{12.9k} & 14.4k \\
\rowcolor{gray!12}
Norm. Avg. & \textbf{0.75} & 0.90 & 0.87 & \underline{0.84} & 0.89 & 1.00 \\
\bottomrule
\end{tabular}
\end{table}

\begin{table}[htbp]
\centering
\small
\caption{MAE on LongBench-v2. Norm.\ Avg.\ is the macro-average of MAE normalized at each prediction point by the history-median MAE. $k$ denotes thousands of tokens. Best and second-best results are in bold and underlined, respectively. Marks are assigned using unrounded values.}
\label{tab:full_longbench}
\setlength{\tabcolsep}{5pt}
\begin{tabular}{lrrrrrr}
\toprule
Prediction point & TokenCast & TRAIL & EGTP & TIE & Self-Pred. & History median \\
\midrule
\multicolumn{7}{l}{\textit{GPT-5.4}} \\
Task Start & \textbf{32.2k} & 37.1k & 36.0k & 35.1k & \underline{32.7k} & 44.0k \\
Call Start & \textbf{53.2} & 60.6 & 56.9 & \underline{53.7} & 61.2 & 74.0 \\
In-call Update & \textbf{35.5} & 73.5 & 68.8 & \underline{62.7} & 70.1 & 91.0 \\
Task Update & \textbf{13.9k} & 26.9k & \underline{24.2k} & 25.1k & 28.2k & 33.0k \\
\rowcolor{gray!12}
Norm. Avg. & \textbf{0.57} & 0.82 & 0.77 & \underline{0.74} & 0.80 & 1.00 \\
\midrule
\multicolumn{7}{l}{\textit{Claude Opus 4.6}} \\
Task Start & \textbf{28.7k} & 33.4k & 32.0k & 30.7k & \underline{29.4k} & 40.5k \\
Call Start & \textbf{48.1} & 57.2 & \underline{50.0} & 54.5 & 56.0 & 67.5 \\
In-call Update & \textbf{41.3} & 69.5 & \underline{57.9} & 61.3 & 65.4 & 82.0 \\
Task Update & \textbf{14.9k} & 23.8k & 22.4k & \underline{21.5k} & 25.1k & 28.5k \\
\rowcolor{gray!12}
Norm. Avg. & \textbf{0.61} & 0.84 & \underline{0.76} & 0.77 & 0.81 & 1.00 \\
\midrule
\multicolumn{7}{l}{\textit{Gemini 3.1 Pro}} \\
Task Start & 39.9k & 43.2k & 41.9k & \textbf{38.2k} & \underline{38.9k} & 50.0k \\
Call Start & \textbf{60.6} & 72.3 & 66.8 & \underline{66.0} & 75.0 & 85.0 \\
In-call Update & \textbf{43.7} & 88.4 & 80.6 & \underline{75.1} & 85.8 & 103.0 \\
Task Update & \textbf{15.8k} & 31.2k & \underline{27.2k} & 28.6k & 33.0k & 38.0k \\
\rowcolor{gray!12}
Norm. Avg. & \textbf{0.59} & 0.85 & 0.78 & \underline{0.76} & 0.84 & 1.00 \\
\midrule
\multicolumn{7}{l}{\textit{DeepSeek-V4-Pro}} \\
Task Start & 52.3k & 55.0k & 54.9k & \underline{51.2k} & \textbf{49.6k} & 61.5k \\
Call Start & \textbf{84.5} & 96.0 & 89.6 & \underline{85.6} & 97.9 & 108.0 \\
In-call Update & \textbf{77.2} & 115.6 & 110.2 & \underline{101.7} & 118.4 & 132.0 \\
Task Update & \textbf{26.1k} & \underline{35.7k} & 38.7k & 37.4k & 42.2k & 47.5k \\
\rowcolor{gray!12}
Norm. Avg. & \textbf{0.69} & 0.85 & 0.84 & \underline{0.80} & 0.87 & 1.00 \\
\midrule
\multicolumn{7}{l}{\textit{Qwen3.8-27B}} \\
Task Start & \textbf{50.8k} & 60.7k & 55.9k & 54.6k & \underline{52.6k} & 70.0k \\
Call Start & \underline{94.4} & 107.8 & \textbf{94.1} & 99.7 & 110.8 & 124.0 \\
In-call Update & \textbf{95.0} & 136.4 & 128.9 & 121.2 & \underline{118.6} & 151.0 \\
Task Update & \textbf{26.9k} & 47.6k & \underline{42.7k} & 44.5k & 49.0k & 56.0k \\
\rowcolor{gray!12}
Norm. Avg. & \textbf{0.65} & 0.87 & \underline{0.79} & 0.80 & 0.83 & 1.00 \\
\midrule
\multicolumn{7}{l}{\textit{Llama-3.2-3B-Instruct}} \\
Task Start & 70.6k & 73.6k & \underline{69.2k} & \textbf{66.3k} & 71.4k & 82.0k \\
Call Start & \underline{116.8} & 128.0 & \textbf{116.2} & 121.8 & 133.7 & 142.0 \\
In-call Update & \textbf{111.9} & 156.2 & \underline{141.9} & 145.2 & 162.7 & 171.0 \\
Task Update & \textbf{38.6k} & 56.8k & 55.4k & \underline{52.1k} & 60.2k & 64.0k \\
\rowcolor{gray!12}
Norm. Avg. & \textbf{0.74} & 0.90 & 0.84 & \underline{0.83} & 0.93 & 1.00 \\
\bottomrule
\end{tabular}
\end{table}

\subsection{Error Relative to Consumption}
\label{app:wape}
Table~\ref{tab:wape} reports WAPE and mean target consumption for GPT-5.4 and Qwen3.8-27B across four benchmarks and four prediction points. For each benchmark, agent LLM, and prediction point, we use the same evaluation weights as MAE:
\begin{equation}
\mathrm{WAPE}(\%)=
100\,\frac{\sum_i w_i|\widehat{y}_i-y_i|}{\sum_i w_i y_i}
=
100\,\frac{\mathrm{MAE}}{\bar{y}},
\qquad
\bar{y}=\sum_i w_i y_i,
\quad
\sum_i w_i=1.
\label{eq:wape}
\end{equation}
Here, $y_i$ is the target consumption: total task consumption $T$ at Task Start, full-call consumption $C_k$ at Call Start and In-call Update, and remaining task consumption $R_k$ at Task Update. The weights follow Appendix~\ref{app:metrics}. All methods within a row share the same target mean and evaluation weights. A WAPE of 15\% means that the weighted MAE is 15\% of the weighted mean target.

\begin{table}[htbp]
\centering
\small
\caption{WAPE (\%) at four prediction points. Best and second-best results are in bold and underlined, respectively. Marks are assigned using unrounded values.}
\label{tab:wape}
\setlength{\tabcolsep}{3pt}
\begin{tabular}{llrrrrrr}
\toprule
Agent LLM & Prediction point & Mean target ($k$) & TokenCast & TRAIL & EGTP & TIE & Self-Pred. \\
\midrule
\multicolumn{8}{c}{\textit{SWE-bench Verified}} \\
\midrule
\multirow{4}{*}{GPT-5.4}
& Task Start & 475.6 & \textbf{30.3} & 34.8 & 33.8 & 33.0 & \underline{32.0} \\
& Call Start & 19.6 & \textbf{0.329} & 0.363 & 0.359 & \underline{0.355} & 0.362 \\
& In-call Update & 20.1 & \textbf{0.194} & 0.393 & \underline{0.371} & 0.385 & 0.399 \\
& Task Update & 286.3 & \textbf{27.9} & \underline{40.2} & 40.9 & 41.6 & 44.0 \\
\midrule
\multirow{4}{*}{Qwen3.8-27B}
& Task Start & 611.5 & \textbf{31.4} & 37.9 & \underline{34.2} & 35.2 & 36.1 \\
& Call Start & 25.4 & \textbf{0.343} & 0.429 & 0.405 & \underline{0.390} & 0.415 \\
& In-call Update & 25.9 & \textbf{0.316} & 0.463 & 0.448 & 0.430 & \underline{0.420} \\
& Task Update & 366.3 & \textbf{33.9} & 48.0 & 46.7 & \underline{44.8} & 47.2 \\
\midrule
\multicolumn{8}{c}{\textit{Search-R1}} \\
\midrule
\multirow{4}{*}{GPT-5.4}
& Task Start & 53.5 & \textbf{63.9} & 70.7 & 70.3 & \underline{68.0} & 69.0 \\
& Call Start & 6.0 & \underline{0.575} & 0.622 & 0.608 & \textbf{0.567} & 0.663 \\
& In-call Update & 6.1 & \textbf{0.513} & 0.646 & 0.631 & \underline{0.603} & 0.654 \\
& Task Update & 31.6 & \textbf{55.7} & \underline{73.1} & 75.0 & 73.7 & 76.6 \\
\midrule
\multirow{4}{*}{Qwen3.8-27B}
& Task Start & 98.1 & \textbf{39.8} & 54.5 & 50.8 & \underline{48.5} & 52.0 \\
& Call Start & 9.5 & \textbf{0.597} & \underline{0.694} & 0.752 & 0.713 & 0.801 \\
& In-call Update & 9.8 & \textbf{0.582} & 0.863 & 0.823 & \underline{0.770} & 0.849 \\
& Task Update & 57.6 & \textbf{41.7} & 68.6 & \underline{59.0} & 61.5 & 66.7 \\
\midrule
\multicolumn{8}{c}{\textit{MMLU-Pro}} \\
\midrule
\multirow{4}{*}{GPT-5.4}
& Task Start & 50.6 & \underline{15.8} & 17.4 & 16.8 & 16.2 & \textbf{14.6} \\
& Call Start & 8.2 & \textbf{0.180} & 0.204 & \underline{0.189} & 0.193 & 0.215 \\
& In-call Update & 8.4 & \textbf{0.110} & 0.233 & 0.218 & \underline{0.199} & 0.207 \\
& Task Update & 27.9 & \textbf{15.4} & 22.2 & 23.3 & 21.5 & \underline{20.8} \\
\midrule
\multirow{4}{*}{Qwen3.8-27B}
& Task Start & 63.8 & \underline{20.4} & 22.4 & \textbf{19.6} & 21.5 & 20.8 \\
& Call Start & 10.0 & \textbf{0.265} & 0.303 & 0.286 & \underline{0.279} & 0.311 \\
& In-call Update & 10.3 & \textbf{0.205} & 0.366 & 0.345 & 0.333 & \underline{0.318} \\
& Task Update & 35.2 & \textbf{23.9} & 31.8 & \underline{29.3} & 30.4 & 32.7 \\
\midrule
\multicolumn{8}{c}{\textit{LongBench-v2}} \\
\midrule
\multirow{4}{*}{GPT-5.4}
& Task Start & 771.2 & \textbf{4.2} & 4.8 & 4.7 & 4.6 & \underline{4.2} \\
& Call Start & 135.7 & \textbf{0.039} & 0.045 & 0.042 & \underline{0.040} & 0.045 \\
& In-call Update & 136.6 & \textbf{0.026} & 0.054 & 0.050 & \underline{0.046} & 0.051 \\
& Task Update & 391.2 & \textbf{3.6} & 6.9 & \underline{6.2} & 6.4 & 7.2 \\
\midrule
\multirow{4}{*}{Qwen3.8-27B}
& Task Start & 856.5 & \textbf{5.9} & 7.1 & 6.5 & 6.4 & \underline{6.1} \\
& Call Start & 112.1 & \underline{0.084} & 0.096 & \textbf{0.084} & 0.089 & 0.099 \\
& In-call Update & 112.6 & \textbf{0.084} & 0.121 & 0.114 & 0.108 & \underline{0.105} \\
& Task Update & 434.2 & \textbf{6.2} & 11.0 & \underline{9.8} & 10.2 & 11.3 \\
\bottomrule
\end{tabular}
\end{table}

\FloatBarrier
\subsection{Generalization}
\label{app:generalization}

\subsubsection{Unseen Task Types}
\label{app:unseen_task_types}

We use the publicly released LiveClawBench trajectories~\citep{long2026liveclawbench}, which contain tasks from 10 application domains executed under a shared agent framework. For each agent LLM, we perform leave-one-domain-out evaluation, treating one domain as the unseen task type and training on the remaining domains. Zero-shot transfer uses no trajectories from the held-out domain. For adaptation, we add $k$ held-out-domain tasks to the training set and evaluate on the remaining held-out tasks. The target-only baseline is trained on the same $k$ tasks without source-domain data. All runs of a task remain in one partition. Relative to Self-Prediction, zero-shot MAE is 1.31 at Call Start and 1.47 at Task Update. With 10 target tasks, the ratios fall to 0.86 and 0.91; with 20, they reach 0.82 and 0.85. The trajectory-selection criteria are in Appendix~\ref{app:collection}.

\subsubsection{Unseen Agent LLMs}
\label{app:unseen_llms}

\begin{figure}[htbp]
\centering
\includegraphics[width=\textwidth]{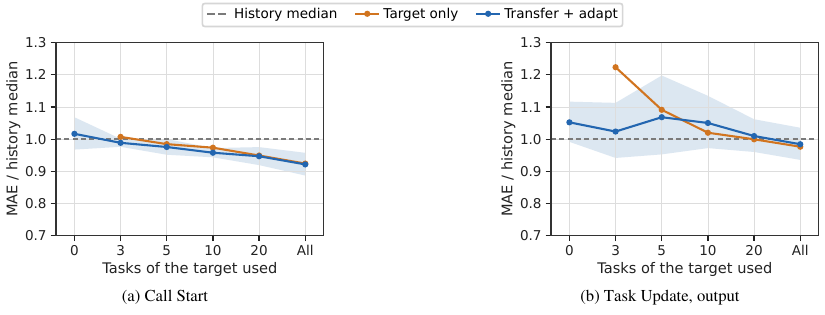}
\caption{Transfer to unseen agent LLMs with increasing amounts of target-model data. The panels report Call Start and remaining output-token prediction at Task Update. MAE is normalized by the target history median; lower is better.}
\label{fig:unseen_llm}
\end{figure}

We hold out each agent LLM in turn and train on trajectories from the remaining models under a shared task-collection and agent framework. Adaptation uses $k$ tasks executed by the held-out model, with task partitions shared across models and adaptation tasks disjoint from the test tasks. At Call Start, zero-shot MAE is 1.02 times the target history median; this ratio falls to 0.96 with 10 target-model tasks and 0.95 with 20. With 3--10 tasks, the transferred predictor achieves lower MAE than training on the same target-model tasks alone. Figure~\ref{fig:unseen_llm} also reports an auxiliary Task Update evaluation of remaining output tokens, a different target from total remaining consumption. Its normalized MAE ranges from 1.01 to 1.07 with up to 20 target-model tasks and reaches 0.98 when all target-model tasks are available. The target-only predictor starts at 1.22 with three tasks and approaches the history median as additional target-model trajectories are provided.

\subsubsection{Cross-Harness Transfer}
\label{app:cross_harness}

Table~\ref{tab:full_harness} reports TokenCast's MAE relative to the history median under each test harness. It also reports transfer from DeepSeek Harness training runs to OpenHands runs of held-out tasks.

\begin{table}[htbp]
\centering
\small
\caption{TokenCast MAE divided by the history-median MAE on the test harness. Lower is better. The two OpenHands rows use the same test runs.}
\label{tab:full_harness}
\setlength{\tabcolsep}{3pt}
\begin{tabular}{llrrrr}
\toprule
Training harness & Test harness & \shortstack{Task\\Start} & \shortstack{Call\\Start} & \shortstack{In-call\\Update} & \shortstack{Task\\Update} \\
\midrule
DeepSeek Harness & DeepSeek Harness & 0.90 & 0.71 & 0.64 & 0.82 \\
OpenHands & OpenHands & 0.83 & 0.77 & 0.68 & 0.75 \\
DeepSeek Harness & OpenHands & 1.04 & 0.88 & 0.79 & 0.96 \\
\bottomrule
\end{tabular}
\end{table}

\subsubsection{Length Extrapolation}
\label{app:length_extrapolation}

We sort the SWE-bench Verified tasks by their median number of calls into five bands, train on the four shorter bands, and evaluate on the longest band. TokenCast achieves lower MAE than Self-Prediction at all four prediction points on these longer tasks, with reductions of 12.4\% at Task Start, 15.7\% at Call Start, 44.9\% at In-call Update, and 26.8\% at Task Update. Table~\ref{tab:length_extrapolation} additionally compares this length-based split with random splits of the same fold sizes. Relative to the random splits, MAE differs by $+7.8\%$ at Task Start, $+1.9\%$ at Call Start, $-2.6\%$ at In-call Update, and $+7.1\%$ at Task Update, so the longer test band raises MAE by at most 7.8\%.

\begin{table}[htbp]
\centering
\small
\caption{Length extrapolation to tasks longer than those observed during training. MAE changes are reported relative to Self-Prediction and matched random splits.}
\label{tab:length_extrapolation}
\begin{tabular}{lrr}
\toprule
Prediction point & vs.\ Self-Prediction $\downarrow$ & vs.\ random split $\downarrow$ \\
\midrule
Task Start & $-12.4\%$ & $+7.8\%$ \\
Call Start & $-15.7\%$ & $+1.9\%$ \\
In-call Update & $-44.9\%$ & $-2.6\%$ \\
Task Update & $-26.8\%$ & $+7.1\%$ \\
\bottomrule
\end{tabular}
\end{table}

\subsubsection{Reasoning Configuration Shift}
\label{app:reasoning_off}

We evaluate a shift in GPT-5.4's reasoning configuration by rerunning the same 69 SWE-bench Verified tasks with reasoning effort set to off and comparing them with the corresponding runs under the low setting. Median run consumption decreases from 273k to 238k tokens. When trained and evaluated within the off configuration, TokenCast reduces MAE relative to Self-Prediction by 20.7\% at Call Start, 50.8\% at In-call Update, and 32.2\% at Task Update (Table~\ref{tab:reasoning_off}). We also transfer the predictor trained under the low configuration directly to the off configuration and adapt it using 20 target-configuration tasks. After adaptation, MAE decreases to 37.9 at Call Start, 23.6 at In-call Update, and 76.0k at Task Update.

\begin{table}[htbp]
\centering
\small
\caption{MAE under a GPT-5.4 reasoning-configuration shift on SWE-bench Verified at the three online prediction points. Results include evaluation within the off configuration, direct transfer from low to off, and adaptation with 20 target-configuration tasks.}
\label{tab:reasoning_off}
\setlength{\tabcolsep}{4pt}
\begin{tabular}{lrrrr}
\toprule
& \multicolumn{2}{c}{Off} & \multicolumn{2}{c}{Low $\rightarrow$ Off} \\
\cmidrule(lr){2-3}\cmidrule(lr){4-5}
Prediction point & TokenCast & Self-Pred. & Direct & 20 tasks \\
\midrule
Call Start & 37.1 & 46.8 & 50.7 & 37.9 \\
In-call Update & 22.5 & 45.7 & 33.0 & 23.6 \\
Task Update & 82.0k & 121.0k & 82.0k & 76.0k \\
\bottomrule
\end{tabular}
\end{table}

\FloatBarrier
\subsection{Online Update Frequency}
\label{app:update_frequency}

We measure the prediction overhead of refreshing the Task Update forecast at different intervals on SWE-bench Verified with GPT-5.4. Starting from the same initial forecast, the 1-, 3-, and 5-call settings refresh the prediction after every 1, 3, or 5 completed calls, respectively, and End makes only the initial forecast. At skipped checkpoints, the latest forecast of final task consumption is retained, the tokens confirmed so far are subtracted, and the remaining estimate is clipped at zero.

Table~\ref{tab:update_frequency} reports the mean number of forecasts and cumulative prediction time per run. Counts include the initial Task Start forecast, and times are reported as mean $\pm$ standard deviation across runs. Updating every call makes 19.7 forecasts and takes $32.8 \pm 30.5$\,ms per run. The three-call and five-call schedules reduce these to 6.9 forecasts and $11.9 \pm 10.4$\,ms, and 4.3 forecasts and $7.7 \pm 6.4$\,ms, respectively. End makes only the initial forecast and takes $2.1 \pm 0.8$\,ms. Every-call updating therefore remains a small fraction of the 129\,s median run time reported in Appendix~\ref{app:collection}.

\begin{table}[htbp]
\centering
\small
\caption{Prediction counts and cumulative overhead at online update intervals on SWE-bench Verified. Predictions/run includes the initial Task Start forecast and later Task Update refreshes.}
\label{tab:update_frequency}
\begin{tabular}{lrr}
\toprule
Update interval & Predictions/run $\downarrow$ & Time/run (ms) $\downarrow$ \\
\midrule
1 call & 19.7 & $32.8 \pm 30.5$ \\
3 calls & 6.9 & $11.9 \pm 10.4$ \\
5 calls & 4.3 & $7.7 \pm 6.4$ \\
End & 1.0 & $2.1 \pm 0.8$ \\
\bottomrule
\end{tabular}
\end{table}

\FloatBarrier
\subsection{Ablation Study}
\label{app:ablation}
\noindent\textbf{Base predictor comparison.}\label{app:base_predictor}
Table~\ref{tab:base_predictor} compares base predictors in the same pipeline. Ridge regression and KNN reach normalized average MAEs of 0.93 and 0.97. Random Forest reaches 0.81, and the MLP reaches 0.78 with 87.2\% interval coverage. The three gradient-boosting methods reach 0.75 or lower, and LightGBM gives the lowest error, 0.69, and the highest coverage, 90.6\%.

The base-predictor comparison reports 0.8\,ms for LightGBM inference, compared with 3.9\,ms for XGBoost and 5.2\,ms for CatBoost. Under every-call updating, the complete pipeline, including feature extraction and composition, makes 19.7 task-level forecasts and takes 32.8\,ms per SWE-bench Verified run on average. We use LightGBM as the base predictor.

\begin{table}[htbp]
\centering
\small
\caption{Base predictor comparison on SWE-bench Verified (GPT-5.4). Norm.\ Avg.\ is the normalized MAE averaged over the four prediction points. 90\% Cov.\ is the empirical coverage of the 90\% prediction interval. Best and second-best results in each column are in bold and underlined, respectively. Coverage is ranked high to low and other metrics low to high.}
\label{tab:base_predictor}
\begin{tabular}{@{}lccc@{}}
\toprule
Predictor & Norm.\ Avg.\ $\downarrow$ & 90\% Cov.\,(\%) & Model time (ms) \\
\midrule
Ridge Regression & 0.93 & 82.4 & \textbf{0.2} \\
KNN & 0.97 & 79.6 & 14.3 \\
Random Forest & 0.81 & 86.8 & 2.7 \\
XGBoost & \underline{0.73} & \underline{89.3} & 3.9 \\
CatBoost & 0.75 & 88.7 & 5.2 \\
MLP & 0.78 & 87.2 & 6.1 \\
\midrule
LightGBM & \textbf{0.69} & \textbf{90.6} & \underline{0.8} \\
\bottomrule
\end{tabular}
\end{table}

\noindent\textbf{Prediction strategy.}\label{app:comp_vs_direct}
At Task Start, no segment has finished, so the prefix--suffix decomposition has no observed anchor. The compositional path has a normalized MAE of 0.87, compared with 0.82 for direct regression. At Task Update, observed segment boundaries provide information about the input baseline and context growth. The compositional path improves to 0.66, below direct regression's 0.72. Averaging the two paths reduces MAE to 0.81 at Task Start and 0.64 at Task Update. The correction model uses the direct and compositional forecasts together with the predicted boundary variables, reducing these values to 0.80 and 0.62, with 90.6\% coverage (Table~\ref{tab:comp_vs_direct}). It is trained on out-of-fold outputs of the complete pipeline.

\begin{table}[htbp]
\centering
\small
\caption{Effect of forecasting strategy on SWE-bench Verified (GPT-5.4). Task Start and Task Update columns report the normalized MAE at these two task-level prediction settings. Best and second-best results in each column are in bold and underlined, respectively. Coverage is ranked high to low and other metrics low to high.}
\label{tab:comp_vs_direct}
\begin{tabular}{@{}lccc@{}}
\toprule
Strategy & Task Start $\downarrow$ & Task Update $\downarrow$ & 90\% Cov.\,(\%) \\
\midrule
Direct only & 0.82 & 0.72 & 88.1 \\
Compositional only & 0.87 & 0.66 & 89.3 \\
Direct--compositional average & \underline{0.81} & \underline{0.64} & \underline{89.8} \\
Full correction pipeline & \textbf{0.80} & \textbf{0.62} & \textbf{90.6} \\
\bottomrule
\end{tabular}
\end{table}

\noindent\textbf{Segment representation ablation.}\label{app:segment_ablation}
Regressing directly on raw execution features raises the normalized average MAE to 0.85, with task-level and call-level MAE of 0.88 and 0.82. Retaining the triple while removing composition gives 0.78. This intervention also changes call-level MAE from 0.68 to 0.74, so the ablation affects more than the task-level composition readout.

Removing $g$ raises the normalized average MAE to 0.76 and task-level MAE from 0.71 to 0.79. Removing $b$ raises it to 0.73 (Table~\ref{tab:segment_ablation}). The larger effect of removing $g$ is consistent with its role in the cross-term $n_Bg_A$; the ablations also change call-level predictions and do not isolate this term.

\begin{table}[htbp]
\centering
\small
\caption{Ablation of the segment representation on SWE-bench Verified (GPT-5.4). Task-level and Call-level columns report normalized MAE averaged over the two prediction points within each level. Best and second-best results in each column are in bold and underlined, respectively.}
\label{tab:segment_ablation}
\begin{tabular}{@{}lccc@{}}
\toprule
Variant & Norm.\ Avg.\ $\downarrow$ & Task-level $\downarrow$ & Call-level $\downarrow$ \\
\midrule
Raw features & 0.85 & 0.88 & 0.82 \\
No composition & 0.78 & 0.81 & 0.74 \\
Drop $g$ & 0.76 & 0.79 & 0.72 \\
Drop $b$ & \underline{0.73} & \underline{0.75} & \underline{0.70} \\
Full & \textbf{0.69} & \textbf{0.71} & \textbf{0.68} \\
\bottomrule
\end{tabular}
\end{table}

\noindent\textbf{Component ablation.}\label{app:component_ablation}
The suffix model takes predictions from the prefix model as input. A shift in this input between training and inference can therefore propagate errors through the cascade. Cross-fitting reduces the shift, while the cost-weighted loss scales errors in the coupled variables $g$ and $n$ by their downstream impact. Removing either component raises the normalized average MAE by 0.03--0.04. Removing both raises MAE to 0.78 and lowers coverage to 87.4\% (Table~\ref{tab:component_ablation}).

Without the correction model, MAE rises to 0.74 and coverage falls to 89.0\%. Removing the boundary variables raises MAE to 0.72, consistent with their role in conditioning the suffix model.

\begin{table}[htbp]
\centering
\small
\caption{Component ablation on SWE-bench Verified (GPT-5.4). Each row removes one component from the full pipeline. Best and second-best results in each column are in bold and underlined, respectively. Coverage is ranked high to low and other metrics low to high.}
\label{tab:component_ablation}
\begin{tabular}{@{}lcc@{}}
\toprule
Configuration & Norm.\ Avg.\ $\downarrow$ & 90\% Cov.\,(\%) \\
\midrule
Full & \textbf{0.69} & \textbf{90.6} \\
$-$ Cross-fitting & 0.73 & 88.9 \\
$-$ Cost weighting & \underline{0.72} & 89.4 \\
$-$ Correction & 0.74 & 89.0 \\
$-$ Boundary vars & \underline{0.72} & \underline{90.1} \\
$-$ Cross-fit.\ \& cost-wt. & 0.78 & 87.4 \\
\bottomrule
\end{tabular}
\end{table}

\FloatBarrier
\subsection{Budget Control}
\label{app:budget_control}

\noindent\textbf{Replay and stopping rules.}
We replay 288 GPT-5.4 runs from 144 SWE-bench Verified tasks, with two recorded runs per task. Seven budgets are defined by the 0.3 to 0.9 quantiles of recorded run consumption. At each Task Update checkpoint, the controller adds confirmed consumption to a selected quantile of predicted remaining consumption and stops the run when the sum exceeds the budget. A replay is trace-complete if it reaches its recorded terminal state. For each method, budget, and test fold, the controller selects the 0.05, 0.5, or 0.95 forecast quantile on runs outside the test fold, choosing the one using the fewest tokens while reaching at least the fixed-budget trace-completion rate. Results are pooled over out-of-fold test runs and averaged over three split seeds.

\begin{figure}[htbp]
\centering
\includegraphics[width=\textwidth]{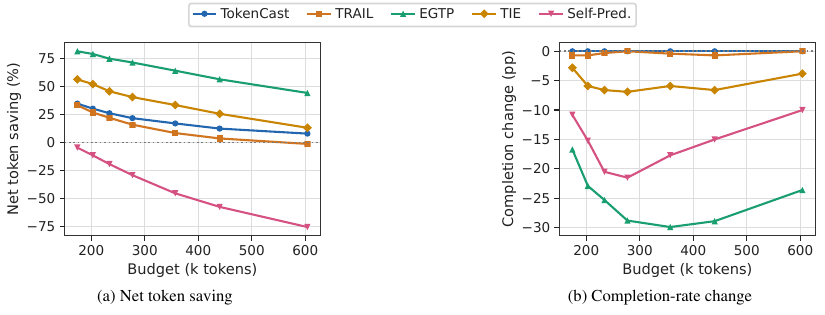}
\caption{Trace completion and token consumption across seven budgets in the offline replay.}
\label{fig:budget_response}
\end{figure}

\noindent\textbf{Cost accounting.}
A controller-stopped run is charged the consumption recorded at its stopping checkpoint. A run that finishes before the fixed cap is charged its recorded total; otherwise it is stopped at the cap, with wall time scaled by the recorded token progress. Prediction processing and time are included in the replay accounting. For encoder-based baselines, prediction tokens count locally processed text; Self-Prediction's count includes additional provider-recorded LLM usage. Savings are relative to the fixed-budget strategy at the same budget. Table~\ref{tab:budget_control} lists the complete results, Figure~\ref{fig:budget_response} shows the budget-wise response, and Table~\ref{tab:prediction_overhead} compares prediction overhead.

\begin{table}[htbp]
\centering
\small
\caption{Budget control on 288 GPT-5.4 runs from 144 SWE-bench Verified tasks. Tokens are in thousands per run, wall time is in seconds per run, and trace completion and savings are in percent. Only TokenCast matches the fixed-budget trace completion at every budget.}
\label{tab:budget_control}
\setlength{\tabcolsep}{4.5pt}
\begin{tabular}{llrrrrrr}
\toprule
\shortstack{Budget\\(k tokens)} & Method & \shortstack{Trace-complete\\(\%)} & \shortstack{Execution\\(k/run)} & \shortstack{Prediction\\(k/run)} & \shortstack{Total\\(k/run)} & \shortstack{Time\\(s/run)} & \shortstack{Saving\\(\%)} \\
\midrule
\multirow{6}{*}{174} & Fixed budget & 30.2 & 154.7 & 0.0 & 154.7 & 113.7 & -- \\
& TokenCast & 30.2 & 101.1 & 0.0 & 101.1 & 76.5 & 34.6 \\
& TRAIL & 29.5 & 91.0 & 12.5 & 103.5 & 73.0 & 33.1 \\
& EGTP & 13.5 & 25.3 & 3.6 & 28.9 & 22.6 & 81.3 \\
& TIE & 27.4 & 59.8 & 8.0 & 67.8 & 45.6 & 56.2 \\
& Self-Pred. & 19.4 & 70.0 & 92.0 & 162.0 & 303.5 & -4.7 \\
\midrule
\multirow{6}{*}{203} & Fixed budget & 39.9 & 173.5 & 0.0 & 173.5 & 125.3 & -- \\
& TokenCast & 39.9 & 121.2 & 0.0 & 121.2 & 90.0 & 30.1 \\
& TRAIL & 39.2 & 112.6 & 14.5 & 127.1 & 87.6 & 26.7 \\
& EGTP & 17.0 & 31.8 & 4.8 & 36.6 & 27.6 & 78.9 \\
& TIE & 34.0 & 73.5 & 9.7 & 83.2 & 54.9 & 52.0 \\
& Self-Pred. & 24.7 & 84.5 & 109.2 & 193.7 & 355.6 & -11.6 \\
\midrule
\multirow{6}{*}{234} & Fixed budget & 50.0 & 190.7 & 0.0 & 190.7 & 135.8 & -- \\
& TokenCast & 50.0 & 141.2 & 0.0 & 141.2 & 103.1 & 26.0 \\
& TRAIL & 49.7 & 132.3 & 16.7 & 149.0 & 101.0 & 21.9 \\
& EGTP & 24.7 & 42.7 & 5.6 & 48.3 & 34.8 & 74.7 \\
& TIE & 43.4 & 92.3 & 11.4 & 103.7 & 67.1 & 45.6 \\
& Self-Pred. & 29.5 & 100.2 & 127.5 & 227.7 & 413.3 & -19.4 \\
\midrule
\multirow{6}{*}{277} & Fixed budget & 59.7 & 210.4 & 0.0 & 210.4 & 148.5 & -- \\
& TokenCast & 59.7 & 165.0 & 0.0 & 165.0 & 118.1 & 21.6 \\
& TRAIL & 59.7 & 157.3 & 19.9 & 177.2 & 118.7 & 15.8 \\
& EGTP & 30.9 & 53.4 & 6.9 & 60.3 & 42.1 & 71.3 \\
& TIE & 52.8 & 111.5 & 13.8 & 125.3 & 80.1 & 40.4 \\
& Self-Pred. & 38.2 & 120.4 & 151.7 & 272.1 & 481.7 & -29.3 \\
\midrule
\multirow{6}{*}{357} & Fixed budget & 69.8 & 237.4 & 0.0 & 237.4 & 165.1 & -- \\
& TokenCast & 69.8 & 197.2 & 0.0 & 197.2 & 138.3 & 16.9 \\
& TRAIL & 69.4 & 193.9 & 23.8 & 217.7 & 144.2 & 8.3 \\
& EGTP & 39.9 & 76.0 & 9.5 & 85.5 & 58.9 & 64.0 \\
& TIE & 63.9 & 141.2 & 17.1 & 158.3 & 98.8 & 33.3 \\
& Self-Pred. & 52.1 & 154.8 & 190.5 & 345.3 & 593.9 & -45.5 \\
\midrule
\multirow{6}{*}{440} & Fixed budget & 79.9 & 258.2 & 0.0 & 258.2 & 177.2 & -- \\
& TokenCast & 79.9 & 226.5 & 0.0 & 226.5 & 156.4 & 12.3 \\
& TRAIL & 79.2 & 222.3 & 26.9 & 249.2 & 163.1 & 3.5 \\
& EGTP & 51.0 & 100.2 & 12.7 & 112.9 & 77.3 & 56.3 \\
& TIE & 73.3 & 171.5 & 20.8 & 192.3 & 119.6 & 25.5 \\
& Self-Pred. & 64.9 & 184.3 & 222.5 & 406.8 & 688.3 & -57.6 \\
\midrule
\multirow{6}{*}{604} & Fixed budget & 89.9 & 280.7 & 0.0 & 280.7 & 191.3 & -- \\
& TokenCast & 89.9 & 258.8 & 0.0 & 258.8 & 176.4 & 7.8 \\
& TRAIL & 89.9 & 254.1 & 30.4 & 284.5 & 184.2 & -1.4 \\
& EGTP & 66.3 & 140.3 & 16.6 & 156.9 & 104.3 & 44.1 \\
& TIE & 86.1 & 218.6 & 25.6 & 244.2 & 150.4 & 13.0 \\
& Self-Pred. & 79.9 & 224.6 & 268.3 & 492.9 & 820.8 & -75.6 \\
\bottomrule
\end{tabular}
\end{table}

\begin{table}[htbp]
\centering
\small
\caption{Budget-equal-weighted replay means. Trace completion is shown alongside overhead because lower processing totals can result from stopping more runs. Encoder-based baselines count locally processed input tokens; Self-Prediction counts additional LLM usage. These token counts therefore represent different resources.}
\label{tab:prediction_overhead}
\begin{tabular}{lrrrr}
\toprule
Method & Trace-complete (\%) & Prediction tokens & Total tokens & Total time \\
\midrule
TokenCast & 59.9 & 0.0k & 173.0k & 122.7\,s \\
TRAIL & 59.5 & 20.7k & 186.9k & 124.5\,s \\
EGTP & 34.8 & 8.5k & 75.6k & 52.5\,s \\
TIE & 54.4 & 15.2k & 139.3k & 88.1\,s \\
Self-Prediction & 44.1 & 166.0k & 300.1k & 522.4\,s \\
\bottomrule
\end{tabular}
\end{table}

\FloatBarrier
\section{Case Studies}
\label{app:case_studies}

Token consumption reflects both the actions needed to finish a task and the context carried into each call. Two executions connect these quantities to observable events: verification after an edit and artifact preparation with a long context. Task Update estimates are displayed as final totals, $\widehat{T}_k=S_k+\widehat{R}_k$, where $S_k$ is confirmed consumption. Call-level estimates retain the full current-call target $C_k$. Dashed lines mark retrospective targets. Grouped bars show Call Start forecasts from TokenCast and Self-Prediction for full current-call consumption.

\FloatBarrier
\subsection{Code Repair}
\label{app:case_code}

GPT-5.4 repairs a \texttt{swap\_dims()} mutation issue in \texttt{pydata\_\_xarray-6938} with DeepSeek Harness. The run contains 21 calls and 39 checkpoints, consumes 243,371 tokens, and lasts 130 s. The following timeline, detail plots, and table show how forecasts change during source editing, verification failure, retry, and completion.

\begin{figure}[htbp]
\centering
\includegraphics{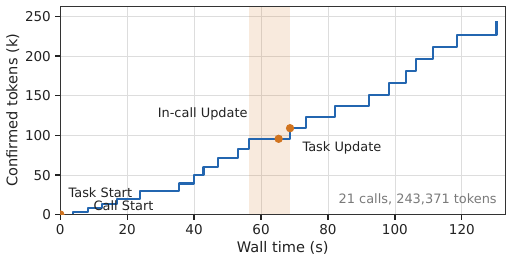}
\caption{Confirmed tokens over wall time for one run. The shaded span is call 12.}
\label{fig:example_run}
\end{figure}

\noindent\textbf{Execution timeline.}
Figure~\ref{fig:example_run} plots confirmed consumption over wall time. Task Start is at 0 s and the first Call Start 1 ms later, when the first request is assembled. The In-call Update shown is checkpoint 5 of call 12 at 65.3 s, with 657 bytes committed and 95,591 tokens confirmed. The Task Update after call 12 is at 68.7 s, with 109,011 tokens confirmed and 134,360 still to come.

\begin{figure}[htbp]
\centering
\includegraphics[width=0.40\textwidth]{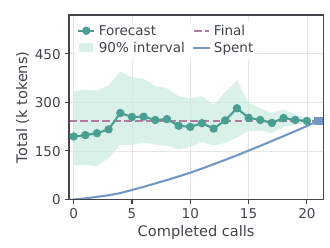}\hfill
\includegraphics[width=0.40\textwidth]{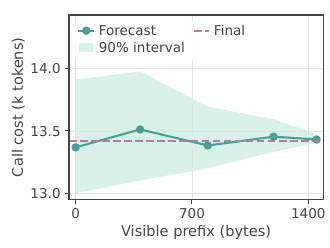}
\caption{Forecast updates during code repair. Left: task-level forecasts following verification outcomes. Right: current-call forecasts during edit call 12 as generation becomes visible.}
\label{fig:case_code}
\end{figure}

\noindent\textbf{Generated prefix.}
Figure~\ref{fig:case_code} and Table~\ref{tab:case_code} report forecasts at selected prediction points. Call 12 performs the source edit and consumes 13,420 tokens. At 795 generated bytes, the prefix exposes the original source span. At 1,450 bytes, it reveals the replacement text. Across these checkpoints, the 90\% interval contracts from 911 tokens at Call Start to 65 tokens at 1,450 bytes, while the forecast remains close to the recorded 13,420-token call cost.

\begin{table}[htbp]
\centering
\caption{Forecasts at selected points in the code-repair execution. Token quantities are in thousands. Recorded targets are shown retrospectively.}
\label{tab:case_code}
\small
\setlength{\tabcolsep}{3pt}
\renewcommand{\arraystretch}{1.13}
\begin{tabularx}{\linewidth}{@{}lXcrr@{}}
\toprule
Prediction point & Available evidence & Target & Forecast [90\% interval] & Recorded \\
\midrule
Task Start & Issue and configuration & $T$ & 194.638 [104.782, 333.912] & 243.371 \\
Call Start, 12 & Edit request assembled & $C_{12}$ & 13.368 [13.001, 13.912] & 13.420 \\
In-call, 12 & 795 bytes, old source span visible & $C_{12}$ & 13.383 [13.201, 13.694] & 13.420 \\
In-call, 12 & 1,450 bytes, replacement text visible & $C_{12}$ & 13.431 [13.407, 13.472] & 13.420 \\
Update, $k=14$ & NumPy alias prevents reproduction & $T$ & 281.746 [191.304, 368.259] & 243.371 \\
Update, $k=15$ & Compatibility workaround, assertions pass & $T$ & 252.381 [211.623, 298.576] & 243.371 \\
\bottomrule
\end{tabularx}
\end{table}

\noindent\textbf{Verification feedback.}
At call 14, reproduction encounters the removed NumPy \texttt{unicode\_alias}, and the predicted task total rises to 281,746 tokens. After a compatibility workaround allows the non-mutation assertions to pass at call 15, the forecast becomes 252,381 tokens, close to the recorded total of 243,371. The execution still consumes another 92,329 tokens while inspecting the diff, preparing artifacts, and submitting the response, accounting for 37.9\% of the final total.

\FloatBarrier
\subsection{Long-Context QA}
\label{app:case_long}

\begin{figure}[htbp]
\centering
\includegraphics[width=0.40\textwidth]{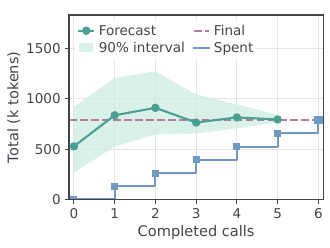}\hfill
\includegraphics[width=0.40\textwidth]{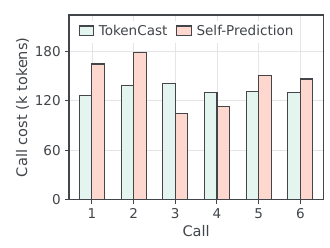}
\caption{Repeated input makes additional calls expensive. Left: task updates after a failed write and artifact preparation. Right: full-call Call Start forecasts from TokenCast and Self-Prediction.}
\label{fig:case_long}
\end{figure}

\noindent\textbf{File operations.}
Qwen3.8-27B answers a LongBench question about OpenLRM and Instant3D using OpenHands. During call 1, it generates option C and its justification, then encounters a file-creation error because \texttt{answer.md} already exists. Inspecting and replacing the placeholder, preparing the patch, and checking the artifacts extend the execution to six calls. Successful replacement at call 3 is followed by another 395,353 tokens of consumption. Figure~\ref{fig:case_long} and Table~\ref{tab:case_long} show the corresponding task-level and call-level forecasts.

\begin{table}[htbp]
\centering
\caption{Long-context stages. All token quantities are in thousands. The same answer is carried through the subsequent file operations.}
\label{tab:case_long}
\small
\setlength{\tabcolsep}{3pt}
\renewcommand{\arraystretch}{1.2}
\begin{tabularx}{\linewidth}{@{}lXrr@{}}
\toprule
Point & Newly available evidence & $S_k$ & Forecast $T$ [90\% interval] \\
\midrule
Task Start & Long-context question and output requirements & 0.000 & 526.384 [264.731, 919.648] \\
Update, $k=1$ & Answer generated, file creation fails & 131.059 & 834.719 [525.193, 1204.762] \\
Update, $k=2$ & Existing answer placeholder inspected & 261.368 & 908.362 [643.881, 1268.997] \\
Update, $k=3$ & Answer file successfully replaced & 392.112 & 762.541 [652.967, 1041.863] \\
Update, $k=5$ & Answer and patch read back & 655.063 & 792.813 [765.208, 839.426] \\
Termination & Finish action recorded & 787.465 & --- \\
\bottomrule
\end{tabularx}
\end{table}

\noindent\textbf{Repeated input.}
The first call processes 129,714 input tokens. Subsequent calls process 130,181--132,209 input tokens and emit 128--350 output tokens each. Input contributes 785,146 of 787,465 tokens (99.7\%). For the recorded execution, the segment decomposition gives
\[
T=nL_1+b
 =6\times129{,}714+9{,}181
 =787{,}465.
\]
The repeated-input term contributes 778,284 tokens. Each additional call therefore carries roughly 130,000 input tokens, so the remaining call count and the input boundary jointly determine the task cost. The failed write at call 1 signals extra calls at this context length, and the Task Update forecast rises from 526k at Task Start to 835k tokens.

\FloatBarrier
\section{Self-Prediction Prompt}
\label{app:self_prediction_prompt}

The following prompt template adapts the zero-shot agent self-prediction protocol of \citet{DBLP:journals/corr/abs-2604-22750} to the four prediction points used in our evaluation. It retains environment inspection, workload analysis, and phase-wise input/output token estimation. At Task Start, the estimator may inspect the initial environment before predicting complete-run consumption. At the other three prediction points, it receives the observed execution record and the target-specific evidence available at that point without advancing the task. Missing fields are marked as \texttt{not available}.

\begin{tcolorbox}[
    colback=black!5!white,
    colframe=black!75!white,
    title={Self-Prediction prompt template},
    fonttitle=\bfseries\footnotesize,
    sharp corners,
    parbox=false,
    breakable
]
\begin{lstlisting}[language={}]
Estimate the token consumption of the agent execution described below. Base your prediction on the task requirements, the specified model and agent configuration, and the evidence available at the supplied prediction point. Your estimate should describe the consumption of the agent operating under its existing workflow and execution limits. Your deliverable is a token-cost estimate. Do not implement a solution, modify task files, or submit an answer to the underlying task.

Prediction target

First, determine which part of the execution the estimate must cover. At Task Start, estimate all input and output tokens that a complete run would consume from the original task state until termination. The estimate includes the exploration, solution development, and verification that the run would require. Any inspection performed specifically to prepare this estimate belongs to the estimation session and is excluded from the predicted consumption. This inspection may improve your understanding of the task; however, the predicted run still begins from its original state.

At Task Update, estimate the input and output tokens that will be consumed after the latest completed call until the task ends. Completed calls provide evidence of progress and consumption, but their recorded usage is outside of this remaining-task target. Their messages and tool results may still appear in later requests; reading that retained content in a later request contributes new input usage within the prediction scope.

At Call Start, estimate the complete input and output consumption of the current call, including retries of its request. At In-call Update, estimate the same complete-call quantity using the generation observed so far. This includes the input, supplied output prefix, continuation, and any usage attributable to retries within the call. A call comprises one initial model request and any retry requests issued before tool feedback is received. A subsequent request made after receiving tool feedback belongs to a later call and is outside the call-level estimate.

A task ends when the agent finishes, fails, or reaches an execution limit. Account for the termination behavior supported by the available evidence. Successful completion is one possible outcome, and repeated failures or exhausted limits may end the run earlier. Execution limits constrain the forecast, but they are not estimates of how much work the agent will perform.

Understanding the task and execution evidence

Read the task statement together with the agent instructions and completion conditions. Determine what the agent must establish, produce, or verify before it can complete the task. Consider how the available tools and workflows organize the work into model calls. Distinguish the number of tool actions from the number of model calls: several actions may be issued in one response, and a single unresolved issue may require multiple responses.

At Task Start, use the available inspection tools to understand the initial environment while leaving task files unchanged. For a coding task, inspect the relevant source files, their dependencies, and the tests associated with the requested behavior. Assess whether the work appears localized or spans several components, whether the expected behavior is clearly specified, and how much investigation is likely to precede a change. For retrieval or document-based tasks, examine the supplied materials and resource descriptions to assess what evidence is already available and what additional information the agent would need to obtain. Focus the inspection on uncertainties that materially affect the expected workload.

At the other prediction points, use the supplied execution record without advancing the task or obtaining additional tool results. Read the actions together with their observed outcomes. A proposed edit, a completed edit, and a successful test provide different evidence of progress. Identify what has been established, what remains unresolved, and which results are still pending. Treat unavailable information as unknown; the absence of a recorded failure does not establish success.

For a task-level forecast, form a plausible continuation from the current state. A coding run may still need to investigate a failure, revise an implementation, run relevant checks, and prepare its submission. A retrieval task may require further searches because the current evidence covers only part of the question. A reasoning task with sufficient information may complete in one response. Use phases that fit the actual task and the observed state. At Task Update, include only phases or portions of phases that remain.

Use the execution history to assess both progress and the cost of further work. Recent calls can indicate typical response lengths, request growth, and the amount of work the agent accomplishes per interaction. Compare calls with similar roles where possible. Repeated searches, recurring test failures, or several calls without resolving an outstanding issue may indicate additional investigation or revision. A sequence of completed checks may indicate that only final verification or submission remains. Relate these observations to the work still required before estimating the remaining call count.

For a call-level forecast, focus on what the current request asks the model to generate. Determine whether the response is likely to contain a short tool invocation, several tool-call arguments, a substantial code fragment, an explanation, or a final answer. At the In-call Update, examine how much of that response has already been produced and what remains unfinished. Use the supplied timing information as supporting evidence when it is informative, while keeping elapsed time distinct from token counts.

Constructing the token estimate

For task-level predictions, the forecast execution is divided into a small number of non-overlapping phases. For each phase, the likely number of model calls, the input those calls will receive, and the output they will generate should be assessed. Let the phase estimates reflect the actual continuation you expect. Final verification or submission may require only one additional call. For the call-level prediction, a single current_call phase is sufficient.

Estimate the input consumption from the content submitted for each model request. This can include system and agent instructions, task statements, tool definitions, retained conversation history, content generated from earlier calls, and tool results. Use an exact supplied input token count when one is available. When later requests have not yet been assembled, estimate their size from the current context, expected additions, and specified context retention behavior.

Account for retained content each time it is submitted. A tool result added early in a run may appear in several subsequent requests; therefore, its contribution depends on both its size and the number of later calls that retain it. Likewise, additional investigation can increase the call count and enlarge the context of those calls. Reflect both effects in the phase estimates. Apply context pruning, summarization, or compaction only when supported by the supplied workflow or observations.

Estimate output consumption from the responses expected within each phase. Include generated text, code, tool-call arguments, and reasoning tokens according to the supplied accounting convention. The final answer may be short, even when intermediate responses consume substantial tokens. When reasoning tokens are already included in the reported output total, count them only once. Tool-generated search results, file contents, and test logs contribute model input when submitted to a later request; their production by a tool is not itself an LLM output.

Use confirmed usage as an accounting anchor. Within the selected scope, include each confirmed input or output count exactly once. At In-call Update, the visible prefix is already part of the complete output being predicted; add only the expected continuation to its counted contribution. Do not add the prefix again when a supplied cumulative output count already includes it. Character and byte counts are observations about text length, not exact token counts; use the supplied token measurements or accounting information where available.

Account for retries within the call according to the request history, observed failures, and the configured retry policy. The recorded usage from retries that have already occurred within the prediction scope is included. Estimate further retry consumption only to the extent supported by the current conditions. The maximum retry allowance defines a limit and does not imply that every retry will occur.

Combine these estimates into the best-supported forecast. Consider whether the evidence supports a straightforward continuation, additional revision cycles, or early termination, and let this assessment inform the predicted workload. Avoid adding unexplained safety margins to each phase.

Preparing the final output

Return a non-negative integer for each token estimate. The phase input estimates must sum to predicted_input_tokens, and the phase output estimates must sum to predicted_output_tokens. Their combined sum must equal the predicted_total_tokens. Check that the result covers the requested scope and includes its confirmed usage without duplication. A complete call estimate must not fall below the confirmed consumption of that call. A Task Update estimate excludes completed call usage and is zero when termination is confirmed and no further model calls remain.

Set predicted_total_tokens to the median (50th percentile) of token consumption for the specified prediction target. Set lower_total_tokens and upper_total_tokens to its 5th and 95th percentiles, respectively, so that the interval targets 90% coverage. These quantiles describe uncertainty in token consumption within the same prediction scope. Return non-negative integers satisfying lower_total_tokens <= predicted_total_tokens <= upper_total_tokens. When confirmed consumption is supplied for that scope, all three estimates must be at least that amount. At Task Update, the scope includes only future consumption; set all three estimates to zero when termination is confirmed and no further model calls remain. Use one breakdown_by_phase entry for each phase included in the forecast, with names that describe the expected work. For either call-level prediction point, use current_call when no further decomposition is needed.

Submit a JSON object with the following fields:

{
  "predicted_input_tokens": <integer>,
  "predicted_output_tokens": <integer>,
  "predicted_total_tokens": <integer>,
  "lower_total_tokens": <integer>,
  "upper_total_tokens": <integer>,
  "breakdown_by_phase": [
    {
      "phase": "<phase name>",
      "input_tokens": <integer>,
      "output_tokens": <integer>
    }
  ]
}

Use the completion interface specified below. Submit the JSON estimate without a task solution or additional explanatory text. Fields marked not available provide no additional evidence and must not be treated as zero-valued observations.

Prediction point:
{{prediction_point}}

Task:
{{task_description}}

Difficulty, when provided:
{{task_difficulty}}

Model and agent configuration:
{{model_and_agent_configuration}}

Agent workflow and available tools:
{{agent_instructions_and_tool_descriptions}}

Execution limits and token accounting:
{{execution_limits_and_token_accounting}}

Initial environment:
{{initial_environment}}

Observed execution history and tool feedback:
{{execution_history_and_tool_feedback}}

Current request:
{{current_request}}

Generated prefix and stream timing:
{{generated_prefix_and_stream_timing}}

Confirmed usage within the prediction scope:
{{confirmed_usage_in_scope}}

Completion instruction:
{{completion_instruction}}

Estimate the consumption for the specified prediction point and submit the JSON estimate.
\end{lstlisting}
\end{tcolorbox}

The \texttt{prediction\_point} field selects one of the four scopes defined above. When the harness exposes a \texttt{finish} tool, \texttt{completion\_instruction} requests submission of the JSON estimate through that tool. Otherwise, it requests the same object as the final response. Estimation calls and inspections performed solely for estimation are recorded as prediction overhead. The input and output fields refer to the selected scope, and therefore, \texttt{predicted\_total\_tokens} represents complete-task consumption at Task Start, complete-call consumption at both call-level points, and remaining-task consumption at Task Update.

\end{document}